%% file: paper.tex
\pdfoutput=1
\documentclass[letterpaper]{article} 
\usepackage[preprint]{aaai2027} 
\usepackage[hyphens]{url} 
\usepackage{graphicx} 
\usepackage{natbib} 
\usepackage{caption} 
\usepackage{algorithm}
\usepackage{algorithmic}
\usepackage{newfloat}
\usepackage{listings}
\DeclareCaptionStyle{ruled}{labelfont=normalfont,labelsep=colon,strut=off}
\floatstyle{ruled}
\newfloat{listing}{tb}{lst}{}
\floatname{listing}{Listing}
\usepackage{booktabs}
\usepackage{placeins}
\usepackage{array}
\renewcommand{\topfraction}{0.95}
\renewcommand{\dbltopfraction}{0.95}
\renewcommand{\textfraction}{0.05}
\renewcommand{\bottomfraction}{0.5}
\renewcommand{\floatpagefraction}{0.85}
\renewcommand{\dblfloatpagefraction}{0.7}
\usepackage{amsmath}
\usepackage{amssymb}
\usepackage{amsthm}

\usepackage{xcolor}
\usepackage{xspace}

\definecolor{cblue}{HTML}{2a78d6}
\definecolor{cred}{HTML}{e34948}
\newcommand{\twin}{\textsc{Twin}\xspace}
\newcommand{\externallink}[2]{%
  \ifdefined\pdfstartlink
    \leavevmode\pdfstartlink attr{/Border[0 0 0]} user{%
      /Subtype/Link/A<</S/URI/URI(#1)>>}%
    \mbox{#2}\pdfendlink
  \else
    \mbox{#2}\ (\url{#1})%
  \fi}
\definecolor{codebg}{gray}{0.955}
\lstdefinestyle{twincode}{
  language=Python,
  backgroundcolor=\color{codebg},
  basicstyle=\scriptsize\ttfamily,
  keywordstyle=\color{cblue}\bfseries,
  commentstyle=\color{cred!70!black}\itshape,
  stringstyle=\color{cblue!60!black},
  frame=single,
  rulecolor=\color{black!18},
  framesep=4pt,
  xleftmargin=2pt,
  xrightmargin=2pt,
  columns=fullflexible,
  breaklines=true,
  breakindent=1em,
  aboveskip=4pt,
  belowskip=5pt
}

\title{\twin: Playing an Unknown Game with a Test-Time Digital Twin}

\author{
Alexy Skoutnev,\textsuperscript{\rm 4}
Kirill Acharya,\textsuperscript{\rm 1}
Gaston Longhitano,\textsuperscript{\rm 3}
\\
Madeleine Udell,\textsuperscript{\rm 1}
Kevin Ellis,\textsuperscript{\rm 2}
Iddo Drori\textsuperscript{\rm 4,1}
}
\affiliations{
\textsuperscript{\rm 1}Stanford University\\
\textsuperscript{\rm 2}Cornell University\\
\textsuperscript{\rm 3}University of Southern California\\
\textsuperscript{\rm 4}Yeshiva University
}

\begin{document}
\maketitle

\begin{abstract}
We present a \textbf{Test-time World-model Inference (\twin)} system, in which a
frontier coding agent writes an executable \textit{world model} for completing
continual learning tasks, such as ARC-AGI-3 games. Traditional approaches
hand-engineer such models, one custom design per task. Each game hides its rules
and goal, and our system constructs them from simulation and interaction alone.
Its inductive prior over grid games is strong enough to recover the true
transitions of the game and the goal on nearly all levels.
Replay validation happens in a twin world model. The harness
enforces that an action is not made until the program reproduces every previous
observed game transition. Each mismatch between a world model prediction and the
actual action result becomes a counterexample that is used to repair the world
model. \twin clears \textbf{179 out of 183 levels (97.8\%)}, and does so more
efficiently than humans in \textbf{158 out of 179 levels (88.3\%)}. The system
infers the goal before any reward on 156 of the levels it clears (87.2\%), and
in the remaining levels automatically discovers the goal by search. The
benchmark scores completion and action efficiency, between 0 and 100,
against humans playing each game for the first time. Played directly, the base
model scores only 7.8\%; an off-the-shelf harness increases it to 61.1\%,
whereas our twin world model increases the same base model to \textbf{93.3\%,
clearing 23 out of 25 games}. Building a usable world model is
simpler than anticipated, whereas the harder problem is inferring the right
goal. The project website replays all 25 runs action by action:
\externallink{https://arc-agi-3-twin.vercel.app/}{\texttt{TWIN website}};
code:
\externallink{https://github.com/Alexyskoutnev/TWIN-ARC-AGI-3}{\texttt{GitHub repository}}.
\end{abstract}

\section{Introduction} 

\begin{figure*}[t]
\centering
\includegraphics[width=1\textwidth]{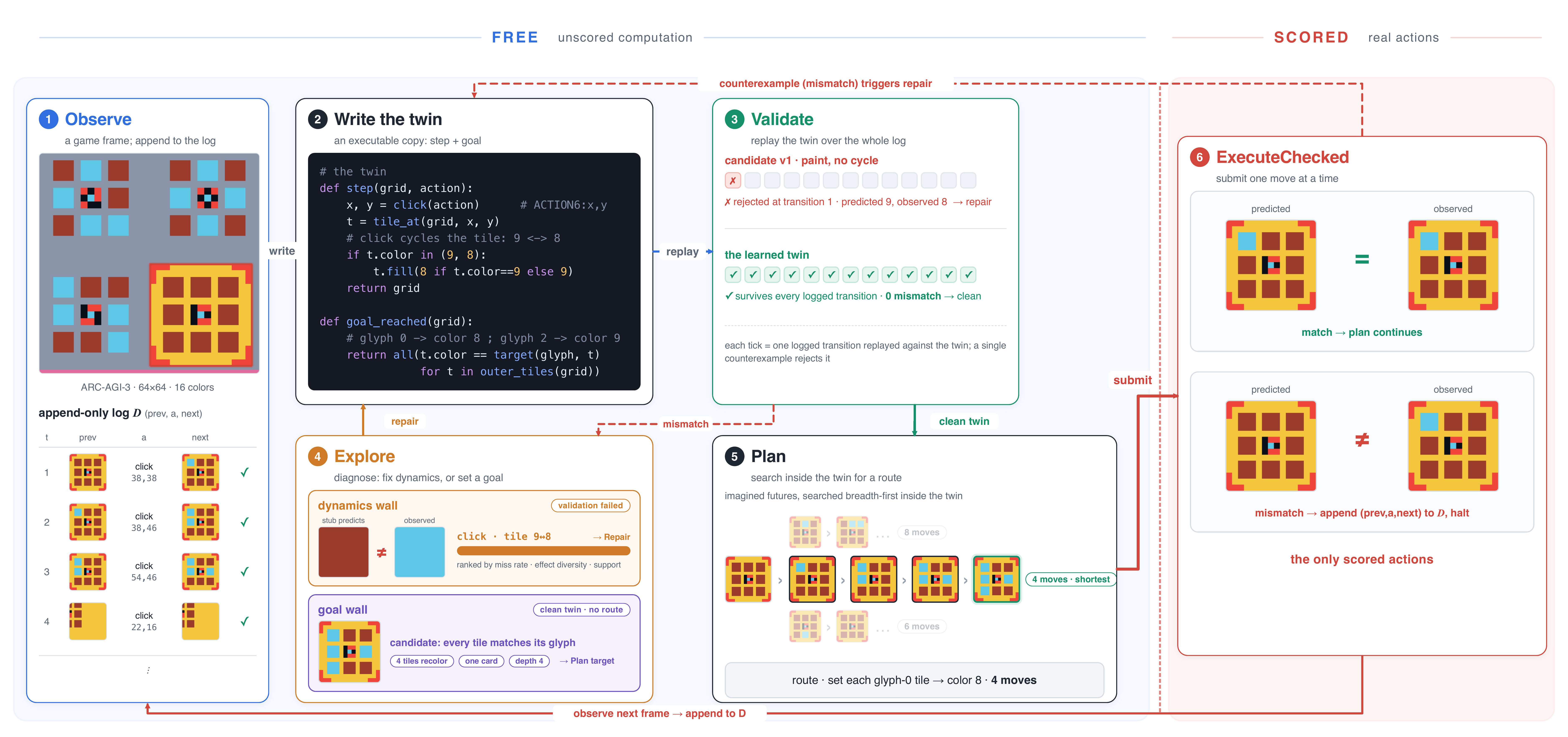}
\caption{The \twin{} loop, illustrated with states from the level ft09. Left of the
scored boundary, every step is unscored. The agent appends each real transition to the log $\mathcal{D}$ and rewrites the twin until \textsc{Validate} replays every logged
transition without a mismatch. \textsc{Explore} then ranks repair targets, or proposes
reachable goal candidates when no route to the goal is found. Planning runs only inside a
validated world model. \textsc{Plan} searches the world model for the shortest route, and
\textsc{ExecuteChecked} submits that route one move at a time against the real ARC-AGI-3 simulator. A halted move returns its transition as a counterexample.}
\label{fig:method}
\end{figure*}

Consider learning an unfamiliar video game without instructions. You press a button and watch what changes. Within a handful of tries, you have a working mental model of the controls, the objects, and what counts as winning. That model lets you plan a few moves and act deliberately rather than mashing buttons. How might an agent acquire and use such a model as efficiently as you do?

We study this on ARC-AGI-3 \citep{arcprize2026arcagi3}, an interactive successor to the ARC reasoning benchmarks \citep{chollet2019measure, chollet2025arcagi2}. Each task is a grid-world game on a $64\times64$ grid of colored cells whose controls and win condition are not stated and must be discovered by playing. Humans completely clear these games, whereas frontier models fail at most games. To close the gap, the field has moved from base models to agentic harnesses, and now to harnesses that build world models of their environment. ARC-AGI-3's action-efficiency score, between 0 and 100, rewards clearing each game in as few actions as a human playing it for the first time. Reasoning, code, and simulation are not scored. Played directly, a strong frontier model scores only 7.8\%, whereas the same model in an off-the-shelf coding harness increases the score to 61.1\%. Introducing a world model into the harness, which builds and validates a model of the game, increases the score to \textbf{93.3\%}. The difficulty is neither knowledge nor perception: these games hide both how the world behaves and what counts as winning.

To act in an unfamiliar world, an agent should build an explicit model of it and reason inside it. \twin{} uses a coding agent to write the world model in code at test time. It treats each game as an unknown deterministic world and builds the twin, a Python program checked against reality, until its predictions match every observed transition. Recovering those rules is the tractable part. Searching inside the twin still needs a target, and the win condition is never stated, so the difficulty remains inferring the goal itself. Goal inference is the exploration problem of reinforcement learning \citep{dearden1998bayesian}: taking actions to discover a goal the environment never states explicitly. Each game therefore poses an \emph{asymmetric objective}: the dynamics can be verified against every interaction, whereas goal reachability must be established through exploration.

Inferring the hidden goal state from sparse feedback is the more challenging problem. \twin{} treats the win condition as a hypothesis to test before any reward exists. Searching inside the validated twin surfaces states that look like progress. The agent writes rival goal predicates over promising planned states and chooses the most efficient plan. Completing a level confirms a candidate, while an exactly reached non-goal eliminates it. The concurrent systems we compare against either wait for the first reward or leave the question unanswered.

\paragraph{Contributions.}
\begin{itemize}\itemsep2pt

\item \textbf{Test-time world-model inference: the agent writes the simulator it plans in.}
The unknown environment becomes an executable hypothesis to inspect, falsify, repair, and plan
through, rather than a black box the agent reacts to. Because the twin must replay every previous
transition before it is trusted, a route is planned and tested before any action is submitted. Of
the actions the agent chose, 92.9\% execute a route already tried in the twin.

\item \textbf{Evaluated on the ARC-AGI-3 benchmark, a world model written at test time is enough
to play an unknown game at human action efficiency.} \twin{} clears 179 of 183 levels and 23
of 25 games, and uses fewer actions than humans playing each game for the first time on 158 of the 179
levels it clears. With the validate--explore--plan loop removed, the same agent clears 148
levels; played directly, the base model clears one game.

\item \textbf{Inferring what counts as winning is the harder half of learning an unknown world.}
A twin recovers how the world behaves from every action it takes, whereas only a completed level
shows what winning looks like. Despite that gap, the first goal \twin{} proposes is the right one
on 156 of the 179 levels it clears, and search finds the goal on the rest.

\end{itemize}

\section{Related Work}
\label{sec:related}

\paragraph{Neural and object-centric world models.} Dyna \citep{sutton1991dyna} established the loop of learning a model, planning through it, and acting to improve it. Modern systems such as World Models \citep{ha2018worldmodels}, MuZero \citep{schrittwieser2020muzero}, and DreamerV3 \citep{hafner2023dreamerv3} learn parametric latent dynamics by gradient descent over many episodes. \twin{} instead induces symbolic source code at test time from tens of transitions, using a pretrained code model \citep{chen2021codex} and execution-guided repair. Code provides exact replay tests, deterministic long-horizon rollouts, and a legible artifact, at the cost of assuming program-compressible, roughly deterministic dynamics. Its object-relative rules mirror learned, object-centric models \citep{kipf2019cswm}; its novelty-driven goal discovery echoes Go-Explore \citep{ecoffet2019goexplore}.

\paragraph{Program induction and LLM agents.} Program induction and test-time adaptation produced the strongest ARC-AGI-1/2 systems \citep{ellis2020dreamcoder,wang2024hypothesis,greenblatt2024arc,li2024induction,chollet2024arcprize}. \twin{} brings this recipe to interaction. Instead of inducing a static input-output transformation, it induces the environment, and its experiments choose its training data. ReAct \citep{yao2023react}, Reflexion \citep{shinn2023reflexion}, Voyager \citep{wang2023voyager}, and Tree-of-Thoughts \citep{yao2023tot} keep the language model in the decision loop. \twin{} instead uses the language model to construct an artifact that removes it from the inner loop; once the twin validates, a multi-step plan runs under machine verification. \twin{} thus spends test-time compute \citep{snell2024testtime} on world-model inference rather than answer sampling; the validated twin fills the verifier role that sample-and-filter \citep{li2022alphacode} and verified-search \citep{hubert2026alphaproof} methods assume.

\paragraph{Related methods tackling ARC-AGI-3.} Five concurrent systems target
the same benchmark. PRO-LONG \citep{fox2026prolong} mandates no world model. Its
contribution is a lossless append-only log the agent queries with code, and
world models emerge in its runs unprompted, one agent reinventing replay
validation on its own. It reports 76.1 pass@1 at its default 500-action
budget and 94.6 at $2{,}000$ actions. PRO-LONG improves what the agent
remembers, whereas \twin{} restricts how it acts: no scored action until a
validated world model explains everything seen. Prime Agent
\citep{primeintellect2026primeagent} takes a different approach: rather than
explicitly modeling the game, it makes the agent's own harness programmable.
It combines recursive language models \citep{zhang2025rlm}, which expose
context and sub-agent calls inside a persistent REPL, with Continual Harness
\citep{karten2026continual}, which lets the agent modify its prompt, skills,
memory, and sub-agents during a run. Prime Agent therefore adapts the agent's
internal machinery, whereas \twin{} builds and repairs an executable model of
the external world. With GPT-5.6 Sol, the base model used by \twin{}, Prime
Agent scores 78.3 and clears 164 of 183 levels. With Claude Opus~5, it
scores 95.5 and clears 179, just past the 95.4 human-expert baseline it
cites. \twin{} reaches 93.3 with Sol. The other three systems converge on one
control loop.
Each logs every interaction, holds the game as an executable program, accepts
the program only when it replays the log, plans inside it, and halts at the
first misprediction. EWM \citep{rodionov2026ewm} established the recipe with a
single coding agent, scoring 63.8. OPINE-World \citep{opine2026world} splits
acting and synthesis across two cooperating agents, steers exploration by a
Bayesian ontology error, and scores 78.4; it declares hidden-state games out
of scope. Schema \citep{zeng2026schema} is the closest design to ours. It
packages the same loop as tools its models drive, backtest and BFS among them,
the loop \twin's harness enforces, and self-reports 98.98 on the public set
through a best-of-two-models fallback per game.

\twin{} differs from all five in two ways. First, before submitting any scored
action, its harness requires the twin to reproduce every transition in the
interaction history. This check is enforced by the harness rather than left to
the agent. The other three world-model systems also test models or plans against
past experience, but they do not require a model to match the entire history
before every real action. Second,
\twin{} hypothesizes the goal before any reward arrives and discriminates rival
goal predicates by the most efficient plan, whereas OPINE-World fits its
predicate only after a first level is cleared and Schema infers
\texttt{is\_goal} alongside its dynamics. The order changes what a scored action
buys: a system without a pre-reward hypothesis spends actions surfacing its
first win, while a correct one converts them into a planned route. We found
that the first goal hypothesis is correct on 156 of 179 completed levels
(87.2\%), so most levels are played goal-directed from the first action.

Other approaches trade explicit world modeling for more search or greater
reliance on the base model. Training-free graph exploration
\citep{rudakov2025graph} reaches deep states through unbounded search, but at
near-zero action efficiency. Direct-play systems instead act without an
explicit model. Claude Opus~5 \citep{anthropic2026opus5card} reports a verified
30.2\% on the semi-private evaluation set, while GPT 5.6 Sol xhigh, the base
model used by \twin{}, scores 7.8\%.

\paragraph{Program-synthesized world models.} WorldCoder introduced the core pattern of using an LLM to synthesize an executable transition and reward model from interaction, plan through it, and repair it from counterexamples \citep{tang2024worldcoder,tang2024rex}. CWM \citep{dainese2024cwm} likewise synthesizes Python dynamics under execution feedback, and PoE-World \citep{piriyakulkij2025poeworld} scales it by composing multiple programmatic experts. \twin{} retains the executable model, replay repair, and model-based planning, but moves them online to raw grid frames with hidden controls and goals. Its central departure is the asymmetric objective. WorldCoder's joint recipe instantiates optimism under uncertainty, the principle behind R-MAX and optimistic MLE \citep{brafman2002rmax,liu2023omle}: among data-consistent models, pursue one that promises reward. \twin{} keeps the optimism but splits the objective. Consistency with every observed transition is a hard precondition for action, whereas optimistic reachability is pursued through planning and goal discovery, not imposed jointly during synthesis.

\section{The \twin{} Method}
\label{sec:method}

Our method has four parts: (i) a problem statement; (ii) three harness routines (validate, explore, plan); (iii) a checked executor; and (iv) goal discovery, as shown in Figure~\ref{fig:method}. \twin{} repeatedly fits an executable world model to a growing interaction log, plans inside it, and executes while checking the prediction at each step. A mismatch returns data for repairing the world model, whereas a consistent model without a goal state results in goal discovery.

\subsection{Problem statement and core representations}

\paragraph{Environment and level protocol.} Each level defines an episodic MDP
$\mathcal{M}=(S,A,T,R,s_0)$. A state $s\in S$ is a rendered $64\times64$ color grid,
and the agent chooses from a menu $A$ of four to six actions, including a click at
grid coordinate $(x,y)$, \texttt{ACTION6:$x,y$}. The transition function
$T:S\times A\to S$ produces the next grid, while the hidden predicate
$R:S\to\{0,1\}$ indicates whether the level is complete. Levels within a game share
an action interface and broad mechanics but change the initial layout. Each game
starts fresh: an identity twin, an empty log, and a new agent context. Twin, log, and agent context persist across the game's levels. At a boundary between levels the agent refactors the twin into rules that replay every logged transition, and cross-level pairs do not enter the log.

\paragraph{Agent interface.} At each step, the agent observes the current grid and the available actions. After acting, it receives the next grid and a completion signal. It is not given object identities, action semantics, the transition or goal rules, demonstrations, or any privileged simulator state. It must infer both how the world changes and what constitutes success solely from interaction.

\paragraph{The executable twin.} A Python file implements
$\hat T:S\times A\to S$ and $\hat R:S\to\{0,1\}$ with a fixed contract:
\begin{center}
\texttt{step(grid, action) -> grid}\qquad
\texttt{goal\_reached(grid) -> bool}
\end{center}
Raw-grid outputs permit cell-by-cell verification, while the implementation may parse objects, update an abstract state, and render back to pixels for more efficient search. The code begins as an identity stub, so every nontrivial rule comes from interaction.

We describe the method by a running example: a public game of cards and tiles (ft09). A click may recolor a tile. Nothing states which cells matter or when a level ends.

\paragraph{Best practices for an uncontaminated run.} The agent plays the game the way a person does. Every scored action passes through the live engine's own interface, five buttons and a click, and nothing else touches game state: no save-scumming, no state teleport, no level skip. Web search is disabled, and an integrity audit scans the transcript and raw agent log for web-tool use or reads of deny-listed ground-truth sources, so a run that cheats is invalidated rather than scored. The harness does not encode game-specific rules, and the public games, released March 2026, are after the base model's GPT 5.6 Sol, February 2026 training cutoff, so every rule in a twin is learned from interaction.

\subsection{The pipeline: validate, explore, plan}

A twin must explain the observed past and support useful imagined futures. Three routines split the work: \textsc{Validate} checks the twin against the log, \textsc{Explore} picks the rule to repair or the destination to try, and \textsc{Plan} searches the validated twin. Because only a submitted real move (\textsc{Submit}) costs score, the loop exhausts the three routines before \textsc{ExecuteChecked} commits any action. Following WorldCoder's consistency and optimism constraints \citep{tang2024worldcoder}, \twin{} enforces the asymmetric objective: fit is required before any scored action, whereas reachability is pursued only through search.

\paragraph{Validate: does the twin explain the past?} Each real move is added
permanently to the transition log $\mathcal D$ as a triple $(s,a,s')$, and the twin must
reproduce every entry:
\begin{equation}
\phi_{\text{fit}}(\mathcal{D},\hat T)\quad\Longleftrightarrow\quad
\forall (s,a,s')\in\mathcal{D},\ \ \hat T(s,a)=s'.
\label{eq:fit}
\end{equation}
\textsc{Validate} is the decision procedure for Eq.~\ref{eq:fit}: it replays the
log and returns either an empty list or the failing transitions and cells, at once a
consistency test and a free bug report. A twin that fails to replay the past is
blocked from acting: no scored move is issued while validation is nonempty. The
check certifies consistency with observed data, not correctness on unseen states. In ft09 the identity stub fails on the first click that recolors a tile, and the mismatched cells return as the first counterexample.

\paragraph{Explore: what blocks progress?} There are only two ways to be stuck: the twin fails to replay the past, the \emph{dynamics wall}, or it replays everything yet reaches no goal, the \emph{goal wall}. \textsc{Explore} reads which wall is live and answers it, one component per half of the asymmetric objective. The \emph{dynamics component} turns validation failures into ranked repair targets for $\hat T$. The \emph{reachability component} turns missing routes into ranked destination candidates. The live wall decides what the next action should teach. 

When validation fails, the dynamics component compiles the failing transitions into a
bug report for the agent, in the spirit of self-debugging from execution feedback
\citep{chen2024selfdebug}. The report groups errors by local context and ranks each
group by how often it fails, how varied its outcomes are, and how little evidence backs
its current rule. Appendix~D specifies the grouping and the ranking signals.
\textsc{Top} sends the highest-priority group and its failing examples to
\textsc{Repair}. The edited $\hat T$ must then replay the full log, preventing a local
fix from breaking an earlier mechanic. In ft09, the report isolates a click that cycles
a tile between two colors; after \textsc{Repair} encodes the cycle, validation passes.

When validation passes but \textsc{Plan} finds no route, the reachability component
searches the twin without $\hat R$, using a larger budget than \textsc{Plan}. It ranks
reachable states by five signs of progress relative to the current frame: a color
appears or disappears, a compact region changes, the scene changes globally, or the
search reaches a new frontier. For each signal, it retains the strongest candidate and
prefers a shorter path when scores tie, because testing costs real actions. The
goal-discovery loop then converts the ranked candidates into hypotheses and tests them
in order.

\paragraph{Plan: what should the agent do if the twin is correct?}
\textsc{Plan} runs breadth-first search with $\hat T$ as the successor function and
$\hat R$ as the goal test, deduplicating full-grid
states. Click games supply a shortlist of candidate coordinates to keep the
branching factor finite. The target is one imagined route to the goal within a
horizon $H$, the planner depth limit:
\begin{equation}
\begin{aligned}
&\phi_{\text{reach}}^{H}(s,\hat T,\hat R)\iff
\exists\, k\in\{0,\ldots,H\},\\
&\exists\, a_1,\ldots,a_k\in A,\ s_0,\ldots,s_k\in S:\\
&s_0=s,\quad s_i=\hat T(s_{i-1},a_i)\ (i=1,\ldots,k),\\
&\hat R(s_k)=1.
\end{aligned}
\label{eq:reach}
\end{equation}
A returned plan satisfies Eq.~\ref{eq:reach} by construction. With unit-cost actions, the first goal found is a shortest plan within the action set, depth limit, and node budget. The budgets are fixed across games: depth 8 with $20{,}000$ nodes, widened to 14 and $30{,}000$ for goal discovery. The quantifiers expose the asymmetry: fit must hold for every logged transition, whereas reachability needs only one imagined route. Returning \textsc{None} means only that the current twin and budget expose no route, which sends control back to \textsc{Explore}. In game ft09, the route is the click sequence that sets each outer tile to the color its center glyph names.

\subsection{Execution turns plans into tests}

For a plan $[a_1,\dots,a_k]$ from state $s_0$, \textsc{ExecuteChecked}
(Algorithm~\ref{alg:twin}) predicts
$\hat s_i=\hat T(\hat s_{i-1},a_i)$, submits actions one at a time, and stops at
\begin{equation}
i^\star = \min\{\, i : s_i \neq \hat s_i \,\},
\label{eq:halt}
\end{equation}
or at a level boundary. A match is verified progress.
A mismatch is appended to $\mathcal D$, invalidates the twin, and blocks further scored
moves until repair. Thus every committed action yields
either progress or one localized counterexample. In game ft09, every click of the final
plan matches its prediction, and the level ends with no wasted action.

\subsection{Goal discovery: learning what counts as success}
\label{sec:goal}

Dynamics and goals receive different supervision. Every action provides a transition
label for $\hat T$, but $\hat R$ receives a positive label only when the environment
signals level completion. At the start of a game, $\hat R$ therefore rejects every
state, leaving \textsc{Plan} without a target. \twin{} resolves this bootstrapping
problem by turning a reachable state into a tentative goal predicate used only to plan
a test. A level boundary confirms the hypothesis; reaching the predicted state without
a boundary rejects it. Interaction thus validates the coding prior's dynamics, while
goal discovery supplies the missing target. At the goal wall, the twin already explains
the observed transitions; what it lacks is a testable destination hypothesis.

\paragraph{Propose and plan.} If validation passes but \textsc{Plan} finds no route,
the goal branch of \textsc{Explore} supplies ranked candidate states. The harness, not
the coding agent, generates one candidate per progress signal and prefers shorter
paths. The coding agent then edits \texttt{goal\_reached} to describe the top
candidate's salient change, producing a tentative $\hat R$. Before spending any scored
action, the harness applies two filters. First, $\hat R$ must evaluate false on every
logged frame; because level completion replaces the winning frame, all logged frames
are known non-goals. This is WorldCoder's observed-data consistency requirement
\citep{tang2024worldcoder}. Second, any candidate previously reached without completing
the level is permanently excluded. A candidate that passes both filters becomes a
temporary target for \textsc{Plan}, which searches the world model for the lowest-cost real test.

\paragraph{Test and update.} The resulting route is executed under the same
halt-on-mismatch guard, producing three distinct outcomes:
\begin{enumerate}\itemsep1pt
\item A level boundary confirms the candidate. The coding agent updates the goal
predicate to accept the predicted pre-boundary candidate state and reject ordinary
logged states, then marks the predicate as confirmed.
\item The candidate state is reached exactly, but no level boundary occurs.
\textsc{RejectCandidate} permanently excludes it, and the next candidate is tested.
\item Reality mismatches $\hat T$ before the candidate is reached. This tests the
dynamics, not the goal: the candidate remains tentative while $\hat T$ is repaired and
the route is replanned.
\end{enumerate}
Only the first two outcomes provide evidence about the goal; a prediction mismatch
provides evidence only about $\hat T$. If \textsc{Explore} finds no reachable candidate,
\textsc{Probe} takes one informative action, either an untried control or an unexplored
click, and appends the resulting transition to the log. Goal discovery therefore
follows a heuristic cycle of proposing, planning, and testing rather than providing a
correctness guarantee. In ft09, the first candidate is a card whose tiles match its
glyph. It triggers a level boundary and is confirmed. Harder games reject several
candidates before finding the goal.

\begin{algorithm}[!t]
\caption{\twin{}, test-time world-model inference. Per game the agent starts an
identity twin and an empty log $\mathcal{D}$. Per move it validates and repairs
the twin, plans a route inside it, and submits under a halt-on-mismatch check.
\textsc{Explore} ranks broken rules at the dynamics wall and reachable candidates
at the goal wall. A proposed goal must evaluate false on every logged frame, and
an exactly reached non-goal is permanently excluded from future proposals. A level
boundary updates the goal, and a mismatch returns a counterexample.}
\label{alg:twin}
\begin{algorithmic}[1]
\STATE $(\hat T,\hat R)\leftarrow$ identity twin;\; $\mathcal{D}\leftarrow\emptyset$ \hfill// per game
\FOR{each level}
 \WHILE{level not complete and budget remains}
 \STATE $s\leftarrow \textsc{Perceive}()$ \hfill// read the real grid
 \WHILE{$\textsc{Validate}(\hat T,\mathcal D)$ returns gaps}
 \STATE $g\leftarrow\textsc{Top}(\textsc{Explore}(s,\mathcal D))$ \hfill// dynamics wall
 \STATE $\hat T\leftarrow\textsc{Repair}(\hat T,g)$
 \ENDWHILE
 \STATE $p\leftarrow\textsc{Plan}(s,\hat T,\hat R)$ \hfill// BFS inside the twin
 \IF{$p=\textsc{None}$}
 \STATE $c\leftarrow\textsc{Top}(\textsc{Explore}(s,\mathcal D))$ \hfill// goal wall
 \STATE $\hat R\leftarrow\textsc{ProposeGoal}(c)$ \hfill// tentative
 \WHILE{$\hat R(s_i)=1$ for some logged frame $s_i$}
 \STATE $c\leftarrow$ next candidate;\; $\hat R\leftarrow\textsc{ProposeGoal}(c)$ \hfill// consistency check, free
 \ENDWHILE
 \STATE $p\leftarrow\textsc{Plan}(s,\hat T,\hat R)$ \hfill// cheapest test of $\hat R$
 \ENDIF
 \IF{$p=\textsc{None}$}
 \STATE $p\leftarrow[\,\textsc{Probe}(s,\mathcal D)\,]$ \hfill// one informative move
 \ENDIF
 \STATE $o\leftarrow\textsc{ExecuteChecked}(p)$ \hfill// halt on mismatch, log to $\mathcal D$
 \IF{$o$ is a level boundary}
 \STATE $\hat R\leftarrow\textsc{UpdateGoal}(o,\mathcal D)$ \hfill// first positive label
 \ELSIF{$o$ reaches $c$ with no boundary}
 \STATE $\textsc{RejectCandidate}(c)$ \hfill// permanently exclude this non-goal
 \ENDIF
 \ENDWHILE
 \ENDFOR
\end{algorithmic}
\end{algorithm}

A twin grows by counterexample-guided refinement, the CEGIS recipe
\citep{solarlezama2008sketching,tang2024rex}: validation
shows that the twin is wrong, exploration shows where, and repair fixes the code.
The agent follows Algorithm~\ref{alg:twin} because it is prompted to. The harness
enforces two replay requirements. No scored move issues until the twin replays
every logged transition (Eq.~\ref{eq:fit}). No scored goal test begins until the
tentative $\hat R$ evaluates false on every logged frame, and a candidate reached
without a level boundary is never proposed again.

\renewcommand{\dbltopfraction}{0.93}
\renewcommand{\dblfloatpagefraction}{0.85}
\section{Experimental Results}

\label{sec:experiments}

We study \twin{} on all 25 public ARC-AGI-3 games. The experiments measure how
well building a digital twin scores, whether every scored action routes through the
validated model, and how far the prior carries before goal discovery takes over. We drive \twin{} with OpenAI Codex running the base model, connected to the game only through files, with no game-specific tools. OPINE-World runs Claude Opus~4.8 and EWM runs GPT-5.5 at high effort, so the Codex ablation and Prime Agent's Sol configuration are the only entries sharing \twin's base model.

\paragraph{Compute.} Across the 25 runs, \twin{} used 2.60 billion processed
tokens and 91.4 hours of wall-clock inference, averaging roughly $224{,}000$ tokens
per scored action. Compute follows difficult goal discovery, not level count: the
ka59 run consumed 24\% of all tokens for less than 10\% of all scored actions.
Per-game usage ranged from 5.1 million to 625 million tokens.

\begin{table}[!b]
\centering\small
\setlength{\tabcolsep}{6pt}
\renewcommand{\arraystretch}{1.12}
\caption{Aggregate performance on the 25-game, 183-level public set. Score averages the benchmark action-efficiency values over games. Won and Levels count full
clears and cleared levels, and all systems use the same published human baselines.
The human row is the score's normalization reference, not a measured system.
\twin{}, as shown in the last row, scores 93.3 of 100 using GPT-5.6 Sol, ahead of OPINE-World 78.4, Prime Agent 78.3 using GPT-5.6 Sol, EWM 63.8 using GPT-5.5, and Codex using GPT-5.6 Sol 61.1.}
\label{tab:agg}
\begin{tabular}{l c c c}
\toprule
System & Score & \shortstack[c]{Won\\(/25)} & \shortstack[c]{Levels\\(/183)} \\
\midrule
Human reference & 100.0 & 25 & 183 \\
OPINE-World & 78.4 & 20 & 160 \\
Prime Agent & 78.3 & -- & 164 \\
EWM & 63.8 & 14 & 146 \\
Codex & 61.1 & 13 & 148 \\
\midrule
\textbf{\twin{}} (ours) & \textbf{93.3} & \textbf{23} & \textbf{179} \\
\bottomrule
\end{tabular}
\end{table}

\begin{figure*}[t]
\centering
\includegraphics[width=0.82\textwidth]{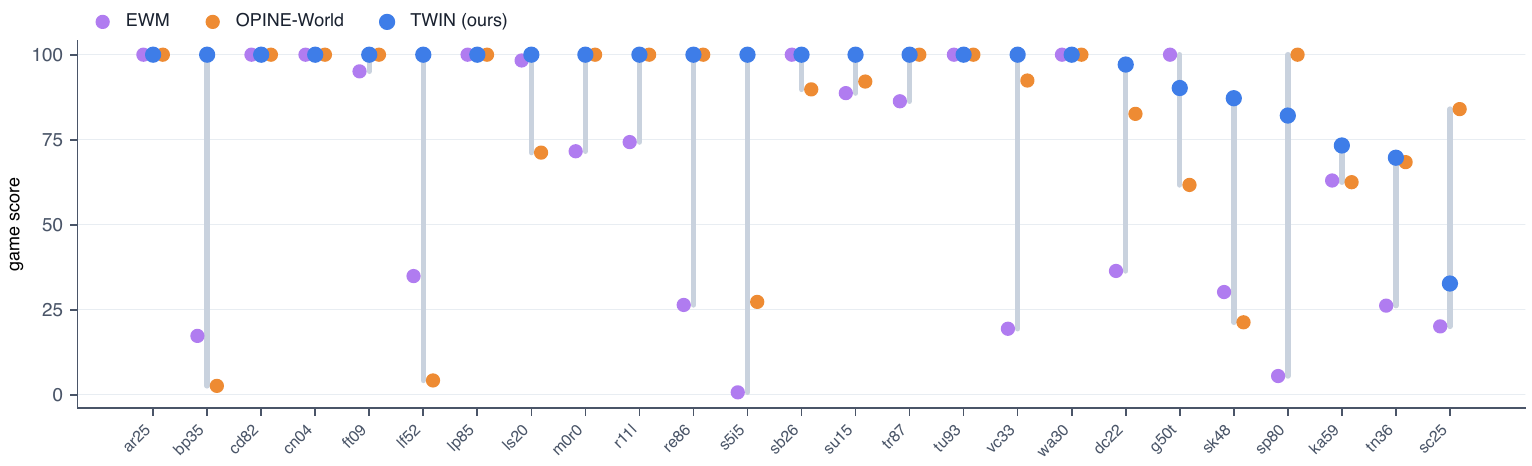}
\caption{Per-game ARC-AGI-3 action-efficiency scores for \twin, OPINE-World,
and EWM, ordered by \twin{} score. For each cleared level $\ell$, the raw efficiency
factor is $e_\ell=\min\{1.15,(h_\ell/a_\ell)^2\}$, where $h_\ell$ and $a_\ell$ are
the human baseline and agent action counts; uncleared levels have $e_\ell=0$. The level-index-weighted game score is capped by the weighted completion fraction, at most 1.0, then scaled to 100. Each game carries three dots joined by a gray range bar, \twin{} blue, OPINE-World orange, EWM violet. The blue dot leads or ties on 22 of the 25, with OPINE-World ahead on g50t, sp80, and sc25.}
\label{fig:scoremap}
\end{figure*}

\begin{figure*}[!t]
\centering
\includegraphics[width=0.60\textwidth]{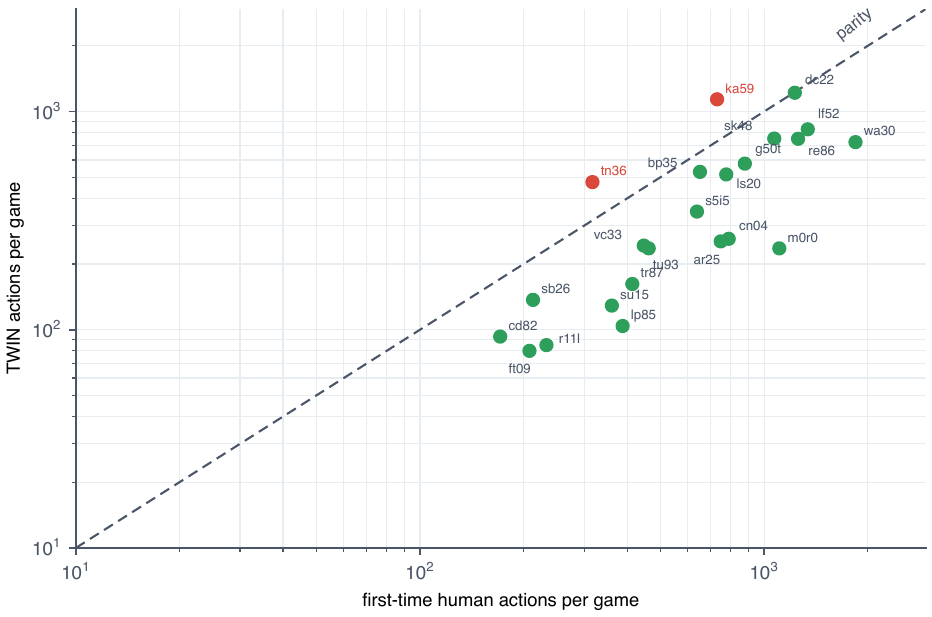}
\caption{Action efficiency relative to the human reference on the 23 games
\twin{} wins. Each dot places one game by its human action count against
\twin's, both axes logarithmic. Green dots lie at or under the dashed
parity line, no more actions than the first-time human, 21 of the 23
at a mean ratio of $0.61\times$. The red dots tn36 ($1.5\times$) and
ka59 ($1.6\times$) lie above.}
\label{fig:parity}
\end{figure*}

\subsection{Building a twin plays ARC-AGI-3 well}

\twin{} reaches a mean score of 93.3 out of 100 on previously unseen games,
clearing 23 of 25 and reaching the 100.0 ceiling on 18
(Table~\ref{tab:agg}, Figure~\ref{fig:scoremap}). OPINE-World scores 78.4, and EWM
scores 63.8 (Table~\ref{tab:main}, Appendix~H). Prime Agent, run on \twin's own
base model, scores 78.3 and clears 164 levels. Played directly, the same base model clears one
game and scores 7.8, so model knowledge alone does not explain the result. The public
games also postdate its training cutoff. The \twin harness accounts for the remaining
85.5 percentage points, and the analyses below identify where that gain appears.

\paragraph{The largest gains appear on the hardest games.} Easy games provide little
separation: all four systems score 100.0 on ar25 and cn04. Differences emerge on
longer multilevel games. \twin{} is the only system to clear bp35, lf52, and sk48,
which contain nine, ten, and eight levels, respectively. On dc22, \twin{} clears all
six levels in $1{,}219$ actions, whereas EWM uses $1{,}842$ actions and clears four.

\paragraph{Harness-enforced validation matters most on long games.} In \twin{},
replay validation is a hard constraint enforced by the harness. The executor blocks
every scored action until the world model reproduces the complete interaction log.
By contrast, the comparison systems rely on prompt instructions to request
validation, which do not mechanically prevent an unvalidated action. This hard
constraint keeps model errors from compounding across long games and may explain
why \twin{} alone clears bp35, lf52, and sk48.

\paragraph{Validated planning reduces scored actions.} To separate efficiency from
coverage, we compare the thirteen games that \twin, EWM, and OPINE-World all fully
clear. Because every system reaches the same endpoint on these games, their action
counts are directly comparable. \twin{} uses $3{,}357$ scored actions, compared
with $5{,}367$ for OPINE-World, $5{,}381$ for EWM, and $7{,}485$ for the human
reference. \twin{} uses the fewest actions on 11 of these 13 games. Thus,
\twin{} solves the same games with fewer interactions, consistent with replay
validation and planning within the world model reducing unnecessary moves. \twin{}
also matches or beats the human action count on 21 of its 23 cleared games and
uses $0.61\times$ as many actions as humans on average
(Figure~\ref{fig:parity}).

\looseness=-1
\paragraph{The worst games invert the comparison.} \twin{} and OPINE-World solve
different subsets of games (Figure~\ref{fig:scoremap}). OPINE-World clears the two
games \twin{} leaves unfinished. sc25 is \twin's worst score, 32.7 against
OPINE-World's 84.0. A hidden countdown on level 4 charges a real test for every
ordering hypothesis, and \twin{} sinks 647 of its 701 actions there. Trial and
error pays no per-hypothesis price. sp80 repeats the shape at 82.1 against
100.0: the twin stays 92.3\% accurate while the sixth level's goal resists, a
goal wall rather than a dynamics wall. The reverse gap is wider. \twin{} reaches
the 100.0 ceiling on bp35 and lf52, where OPINE-World scores 2.6 and 4.2,
stalling by level three of nine and four of ten.

\subsection{The harness earns the score: an ablation}

To test whether the gains come from the method rather than the base agent, we run
off-the-shelf Codex (the Codex columns of Table~\ref{tab:main}). The ablation keeps the identical agent, base model, file bridge, and
sandbox, with the entire \twin{} harness disabled. The deletion removes the executable world model, replay validation
(\textsc{Validate}), diagnostic exploration (\textsc{Explore}), model-based planning
(\textsc{Plan}), and the halt-on-mismatch executor. This comparison isolates the full
harness with the base model fixed. Disabling the \twin{} harness
reduces the mean score from 93.3 to 61.1 and the number of fully cleared games
from 23 to 13 (Table~\ref{tab:agg}). \twin{} matches or exceeds the ablation's
score on 24 of 25 games. The sole exception is sc25, where \twin{} scores
32.7 and the ablation scores 44.8. Under the countdown above, the ablation
clears five levels by trial and error, which scores higher because uncleared
levels receive zero credit.

\paragraph{Every scored action starts from a replay-validated twin.} Replay
validation certifies all logged transitions but does not guarantee predictions in
unseen situations. Of the scored actions, 92.9\% execute plans tested in
simulation, while 7.1\% are deliberate probes chosen to improve the model.
Outcomes disagree with the twin on 20.1\% of actions, and each mismatch becomes a
counterexample for repair. Repairs persist: 31 previously mispredicted situations
recur later, and the repaired twin predicts all 31 correctly.

\paragraph{Dynamics generalize, while goal errors dominate cost.} A pair is
first-seen when no identical state--action pair appears earlier in the log. Each
predicted next frame is hashed and committed before the outcome is observed, so it
cannot be revised after seeing the result. The twin predicts $8{,}210$ of $10{,}392$ first-seen pairs exactly
(79.0\%), compared with 602 of 635 recurring pairs (94.8\%). Novelty
therefore costs 15.8 percentage points, yet nearly four of five unseen pairs remain
cell-exact. Goal errors are less frequent but more costly: the first committed goal
hypothesis is correct on 156 of 179 completed levels (87.2\%). sp80 stalls
goal-limited at 92.3\% dynamics accuracy, while sc25 is mixed at 83.7\% with
costly timer-driven tests. Goal search also raises action counts on tn36 ($1.5\times$ human) and ka59
($1.6\times$). The cost is concentrated: the 23 non-optimal levels add $1{,}291$
actions over the human baselines, and five levels account for 71\%. Goal proposal
and temporal state modeling are therefore the main targets for improvement.

\subsection{Limitations}

\looseness=-1
Replay validation assumes deterministic dynamics representable by the twin and certifies only logged transitions, a boundary a probabilistic twin would loosen. Planning and goal discovery run under fixed search budgets, so a goal beyond the horizon goes unfound. \texttt{step} reads one frame, so mechanics driven by long temporal context, history the grid does not show, are beyond this scope. Yet bp35 hides most of each level behind a scrolling camera, and \twin{} clears all 9 levels under the human count by buying the hidden map one probe per place (Appendix~G), so this class of partial observability is handled too. Truly latent state, a variable no frame ever shows, stays out of scope. Exact replay also presumes small discrete states: cell equality on $64\times64$ grids is free and decidable, while continuous observations would make every replay an approximate comparison with a threshold to tune. Extending \twin{} to probabilistic twins remains future work.

\section{Conclusion}
\enlargethispage{8.8\baselineskip}

\twin{} treats interaction with an unknown game as the problem of constructing and validating an executable world model. A coding agent expresses its current theory as a Python program, checks it against every observed transition, plans inside it, and executes under a halt-on-mismatch guard. Each submitted action therefore becomes planned progress, a concrete counterexample for model repair, or a deliberate probe.

\twin{} scores 93.3 of 100 on the benchmark's action-efficiency metric,
clearing 23 of 25 public games (179 of 183 levels). The same off-the-shelf
Codex without the validate-explore-plan loop scores 61.1 and acts without
predictions validated against the interaction history. Twins predict 79\% of
first-seen state--action pairs exactly, showing reusable rules rather than
transition replay. The first goal hypothesis is correct before any reward on
87.2\% of completed levels (156 of 179); search covers the rest.

\looseness=-1
The unfinished games mark the frontier: accurate dynamics, unresolved goals.
\twin{} demonstrates a new paradigm in which the agent writes the environment as an
executable hypothesis to inspect, falsify, repair, and plan through. Stronger priors
may help agents infer goals as quickly as humans do when playing an unfamiliar
game for the first time, drawing on a lifetime of experience.

\FloatBarrier
\clearpage
\raggedbottom
\bibliography{refs}

\clearpage
\raggedbottom
\section{Appendix A: World-Model Accuracy on Unseen Transitions}

The \twin harness requires the world model to reproduce every transition in
the recorded interaction history before issuing a scored action. Replay
establishes consistency with observed transitions, but a model that merely
memorizes them would also pass. We therefore evaluate each final twin on
previously unobserved state--action pairs. A pair is previously unobserved
when the agent visited its state during the run but never took its action
from that state. For each pair, we give the
same current frame and action to the final twin and the ARC-AGI-3 engine, then
compare the two resulting next frames cell by cell.

To run these comparisons from the states \twin{} visited, we first
reset the ARC-AGI-3 engine and replay the run's recorded actions in order.
Each frame the engine returns must match, cell by cell, the frame recorded
during \twin's original run. The replay is exact for 22 of the 25 games.
For sc25, dc22, and tn36 it diverges partway, so we keep only the prefix
ending at the last matching frame. From each verified trajectory or prefix, we
sample up to 100 visited states. At every sampled state, we test each
non-click action never taken from that state. We also test up to eight click
actions taken elsewhere in the same run but never from that state. Each
comparison of the twin's next frame with the engine's next frame counts as one
branch evaluation, yielding $20{,}790$ evaluations across all 25 games.
sp80's twin stores a call counter, so each of its probes is graded under
the counter state a log-order replay produces at that state; this makes
its grading order-independent, and sp80 joins every measurement below.

\begin{figure*}[p]
\centering
\includegraphics[width=0.88\textwidth]{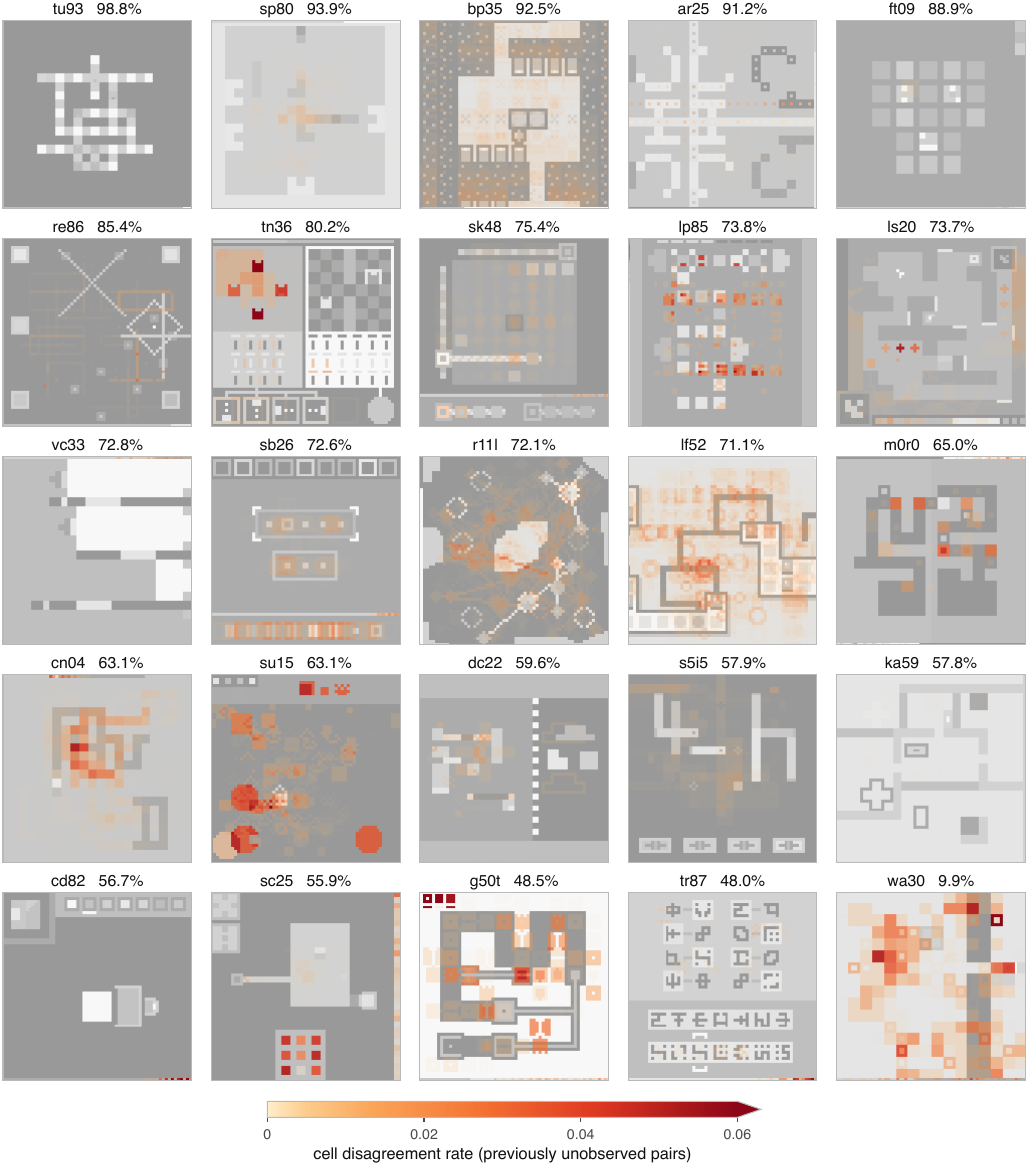}
\caption{Spatial distribution of final-twin errors on previously unobserved
pairs. Each panel shows, in grey, one visited state from that game. Red
shows, per cell on a shared scale, how often the twin's prediction is wrong
there across the game's previously unobserved pairs. Panels are ordered by
the exact-frame accuracy shown in their titles. Most errors are spatially
concentrated: although cd82 misses 43.3\% of its next frames on unobserved
pairs, only 11 of $4{,}096$ cells account for 90\% of its total
disagreement, all within the one-cell progress bar along the bottom edge.
The hottest 5\% of cells contain a median of 60\% of
each twin's disagreement mass. wa30 is the diffuse exception, with errors
spread across the playfield and no row above 2.5\% of the mass. tu93,
ar25, and ft09 remain nearly error-free. Tests begin only from visited
states, and rates include only tests where the twin returns a frame.}
\label{fig:wmheat}
\end{figure*}

The final twins predict the complete next frame exactly for 70.1\% of
previously unobserved pairs. The test measures
generalization beyond replay, which checks only pairs already recorded in the
interaction history. Across games, the exact-match rate ranges from 9.9\%
to 98.8\% (Figure~\ref{fig:wmacc}). tu93, sp80, bp35, and ar25 exceed 91\%,
indicating reusable rules, whereas wa30 reproduces its recorded interaction
history but reaches only 9.9\% on previously unobserved pairs. tu93, lp85, and
cd82 improve on their acting-time accuracy because later repairs also correct
rules needed at earlier states. Most residual errors are spatially localized
to a sprite or counter: in the median game, 561 of $4{,}096$ cells contain
90\% of the disagreement mass (Figure~\ref{fig:wmheat}).

Two limitations qualify this result. First, most cells are unchanged
background, which inflates the cell-match fraction even when an important
object is wrong. We therefore rely on exact-frame accuracy. Second, every test
starts from a visited state and uses an action type or click coordinate
observed in that run. The experiment therefore measures generalization near
the agent's experience, not on arbitrary states or actions.

\begin{table}[H]
\centering\footnotesize
\setlength{\tabcolsep}{4.6pt}
\renewcommand{\arraystretch}{1.08}
\caption{Five next frame comparisons pooled over all 25 games. Each game's
twin is the world model the agent learned while playing that game. Every test
gives the twin one state--action pair and grades the twin's predicted next
frame against a reference. The rows differ in which pairs they use and which reference grades
them. \emph{Tests} counts the comparisons. \emph{Replay} uses the pairs from
the original run, graded against the next frames recorded in that run.
\emph{Observed-pair engine match} grades those same pairs against a fresh
rerun of the ARC-AGI-3 engine. \emph{Unobserved-pair engine match}, the
primary generalization test, grades previously unobserved pairs against the
rerun engine. A previously unobserved pair is a visited state with an action
never taken from that state. sp80's probes are graded under the counter
state a log-order replay produces, since its twin stores a call counter.
\emph{Boundary probes}
grade 98 cross-level or cross-game pairs with no recorded next frames
against the rerun engine. \emph{Exact frames} counts a prediction as exact
only when it agrees with the reference on every one of the
$64\times64 = 4{,}096$ cells. One differing cell fails the whole frame.
\emph{Cell match} is the mean fraction of agreeing cells. The twins exactly
predict 70.1\% of previously unobserved pairs.
Figure~\ref{fig:wmacc} reports results by game.}
\label{tab:beyondlog}
\begin{tabular}{l r r r}
\toprule
Measurement & Tests & \shortstack{Exact\\frames} & \shortstack{Cell\\match} \\
\midrule
Replay of recorded transitions & 11,314 & 99.4\% & 0.9996 \\
Engine match, observed pairs & 2,354 & 98.9\% & 0.9985 \\
Engine match, unobserved pairs & 20,790 & 70.1\% & 0.9956 \\
Cross-level or cross-game probes & 98 & 16.3\% & 0.9014 \\
\bottomrule
\end{tabular}
\end{table}

\begin{figure*}[p]
\centering
\includegraphics[width=0.66\textwidth]{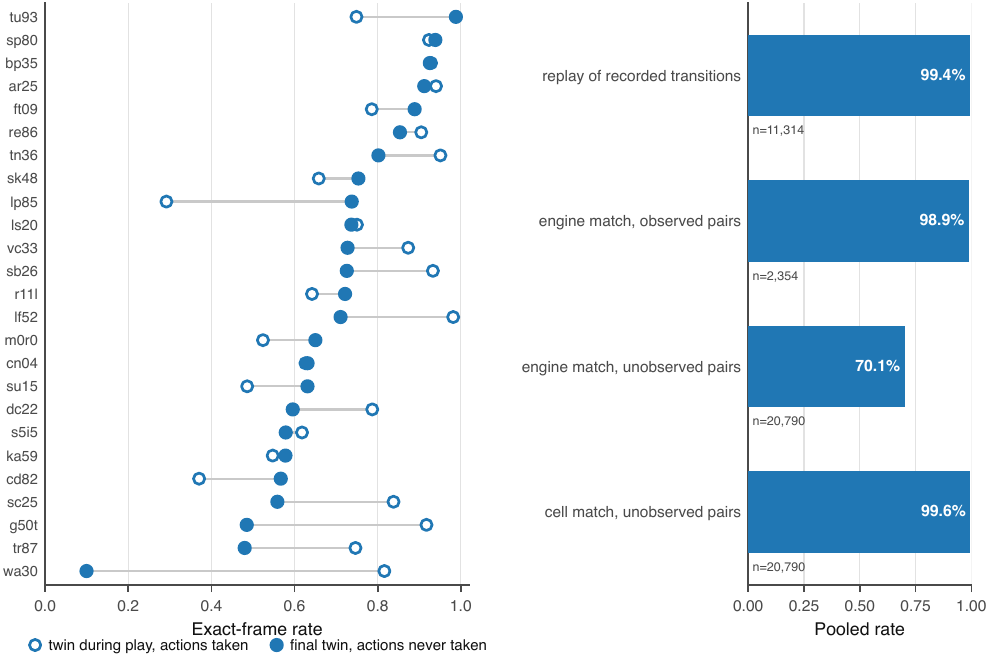}
\caption{Per-game world-model accuracy against the ARC-AGI-3 engine. Every
point compares next frames predicted by a game's twin with next frames the
engine returned. A prediction counts only when all $4{,}096$ cells
match. Left: open circles show how often the twin's predictions were exact during
the run, one prediction before every action. Filled circles show how often
the final twin is exact on pairs it never saw, a visited state with an
action never taken from it. A filled circle to the right of its open circle means the
twin improved during the run. The open circle averages the whole run, early
mistakes included. The filled circle tests only the final, fully repaired
twin. Games are ordered by the filled-circle rate. Right: the same
measurements pooled, with the number of pairs under each bar. The values
correspond to Table~\ref{tab:beyondlog}. Filled-circle accuracy ranges from
9.9\% on wa30 to 98.8\% on tu93. wa30 replays its recorded run almost
perfectly yet fails on pairs it never took: exact replay does not guarantee
generalization.}
\label{fig:wmacc}

\centering\scriptsize
\setlength{\tabcolsep}{9pt}
\renewcommand{\arraystretch}{0.86}
\captionof{table}{Per-game replay and generalization accuracy for each final
twin. \emph{Replay} counts the recorded transitions the final twin reproduces
exactly, over the number tested. \emph{Unobserved-pair exact} grades the final
twin against a rerun of the ARC-AGI-3 engine on pairs it never saw, a visited
state with an action never taken from it. It reports the percentage predicted
exactly, all $4{,}096$ cells matching, and is the per-game version of
Table~\ref{tab:beyondlog}'s primary generalization test. \emph{Cell match} is
the mean fraction of agreeing cells over those same pairs. \emph{Tests} counts
them. Accuracy ranges from 9.9\% on wa30 to 98.8\% on tu93. sp80 is
graded under the counter state a log-order replay produces, since its twin
stores a call counter.}
\label{tab:beyondlog-pergame}
\begin{tabular}{l r r r r}
\toprule
Game & Replay & \shortstack{Unobserved\\pair exact} & \shortstack{Cell\\match} & Tests \\
\midrule
ar25 & 246/246 & 91.2\% & 0.9994 & 1,205 \\
bp35 & 506/506 & 92.5\% & 0.9940 & 939 \\
cd82 & 87/87 & 56.7\% & 0.9999 & 1,145 \\
cn04 & 191/255 & 63.1\% & 0.9961 & 1,215 \\
dc22 & 1,211/1,211 & 59.6\% & 0.9991 & 1,140 \\
ft09 & 74/74 & 88.9\% & 0.9999 & 639 \\
g50t & 570/570 & 48.5\% & 0.9830 & 400 \\
ka59 & 1,123/1,123 & 57.8\% & 0.9998 & 1,107 \\
lf52 & 819/819 & 71.1\% & 0.9917 & 1,151 \\
lp85 & 96/96 & 73.8\% & 0.9981 & 800 \\
ls20 & 508/508 & 73.7\% & 0.9985 & 300 \\
m0r0 & 230/230 & 65.0\% & 0.9987 & 1,206 \\
r11l & 76/79 & 72.1\% & 0.9517 & 675 \\
re86 & 740/740 & 85.4\% & 0.9969 & 396 \\
s5i5 & 340/340 & 57.9\% & 0.9985 & 800 \\
sb26 & 129/129 & 72.6\% & 0.9981 & 893 \\
sc25 & 665/665 & 55.9\% & 0.9987 & 1,126 \\
sk48 & 744/744 & 75.4\% & 0.9984 & 1,112 \\
sp80 & 1,001/1,001 & 93.9\% & 0.9992 & 1,207 \\
su15 & 120/120 & 63.1\% & 0.9848 & 797 \\
tn36 & 447/447 & 80.2\% & 0.9964 & 792 \\
tr87 & 156/156 & 48.0\% & 0.9995 & 300 \\
tu93 & 219/219 & 98.8\% & 1.0000 & 253 \\
vc33 & 236/236 & 72.8\% & 0.9999 & 800 \\
wa30 & 712/713 & 9.9\% & 0.9917 & 392 \\
\bottomrule
\end{tabular}
\end{figure*}

\clearpage

\section{Appendix B: Planning inside the Twin versus the Real Engine}

\begin{figure}[t]
\centering
\includegraphics[width=\columnwidth]{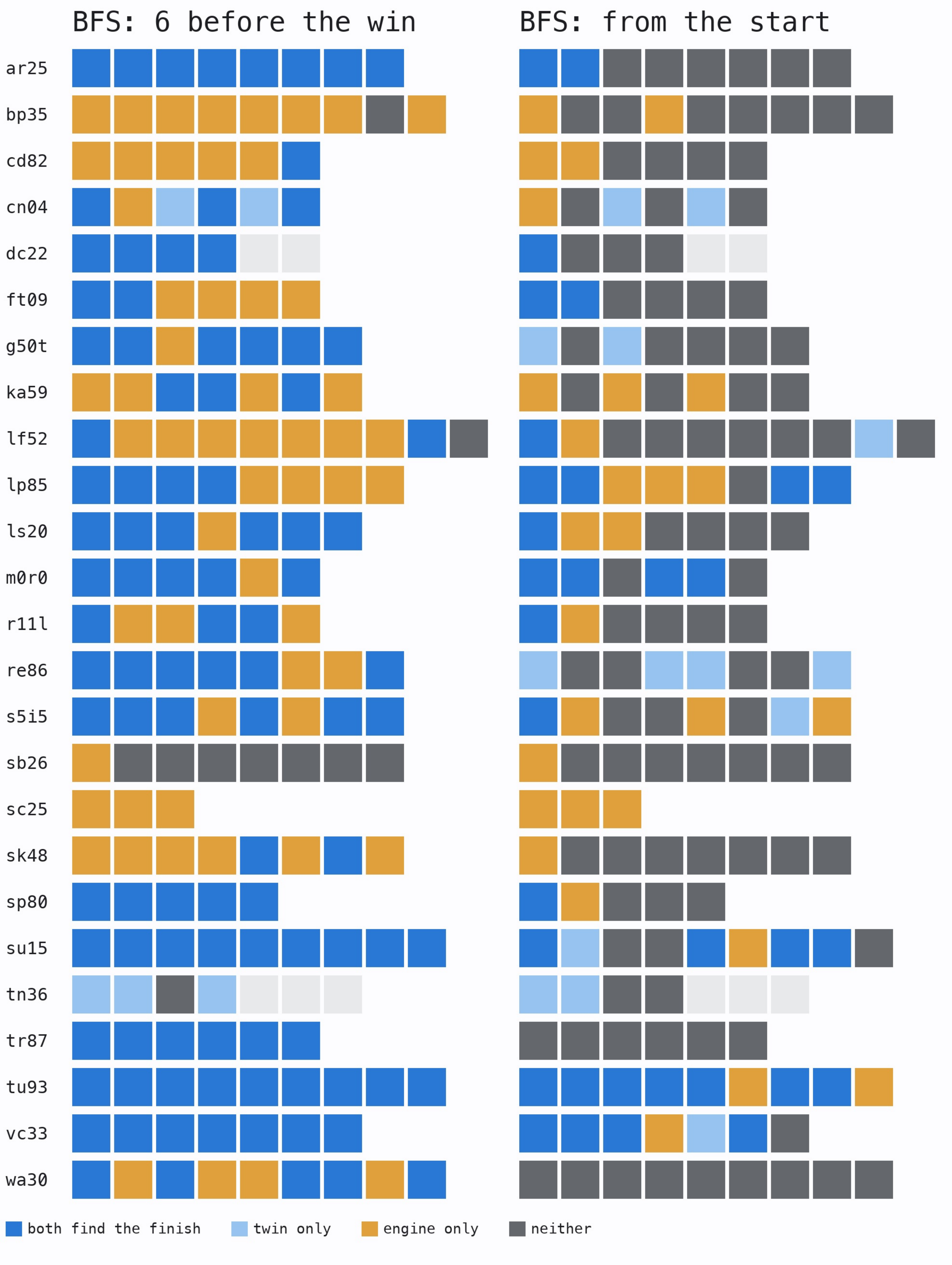}
\caption{We test the same bounded BFS in two worlds, once inside the
twin and once inside the ARC-AGI-3 engine, from the same restored
start state, over all 179 completed levels. The race compares which
model's forward dynamics has a better search to the finishing goal state, the
twin's or the ARC-AGI-3 game engine's own.
One row per game, one
cell per level, in play order. The left block, \emph{BFS: 6 before the
win}, branches six recorded actions before each level's win; the right
block, \emph{BFS: from the start}, branches at the level's entry state.
Blue: both searches find a finish. Orange: only the engine's does. Light
blue: only the twin's search returns a plan. Dark gray:
neither search succeeds within budget. Budgets: $20{,}000$ states per
search in the left block, $200{,}000$ in the right, wall clocks as
safety caps only. Six actions out
the searches broadly agree,
both finishing on 102 levels; from the start the engine finds 61 and
the twin 47, agreeing on 33.}
\label{fig:paritymap}
\end{figure}

\begin{figure*}[p]
\centering
\includegraphics[width=0.70\textwidth]{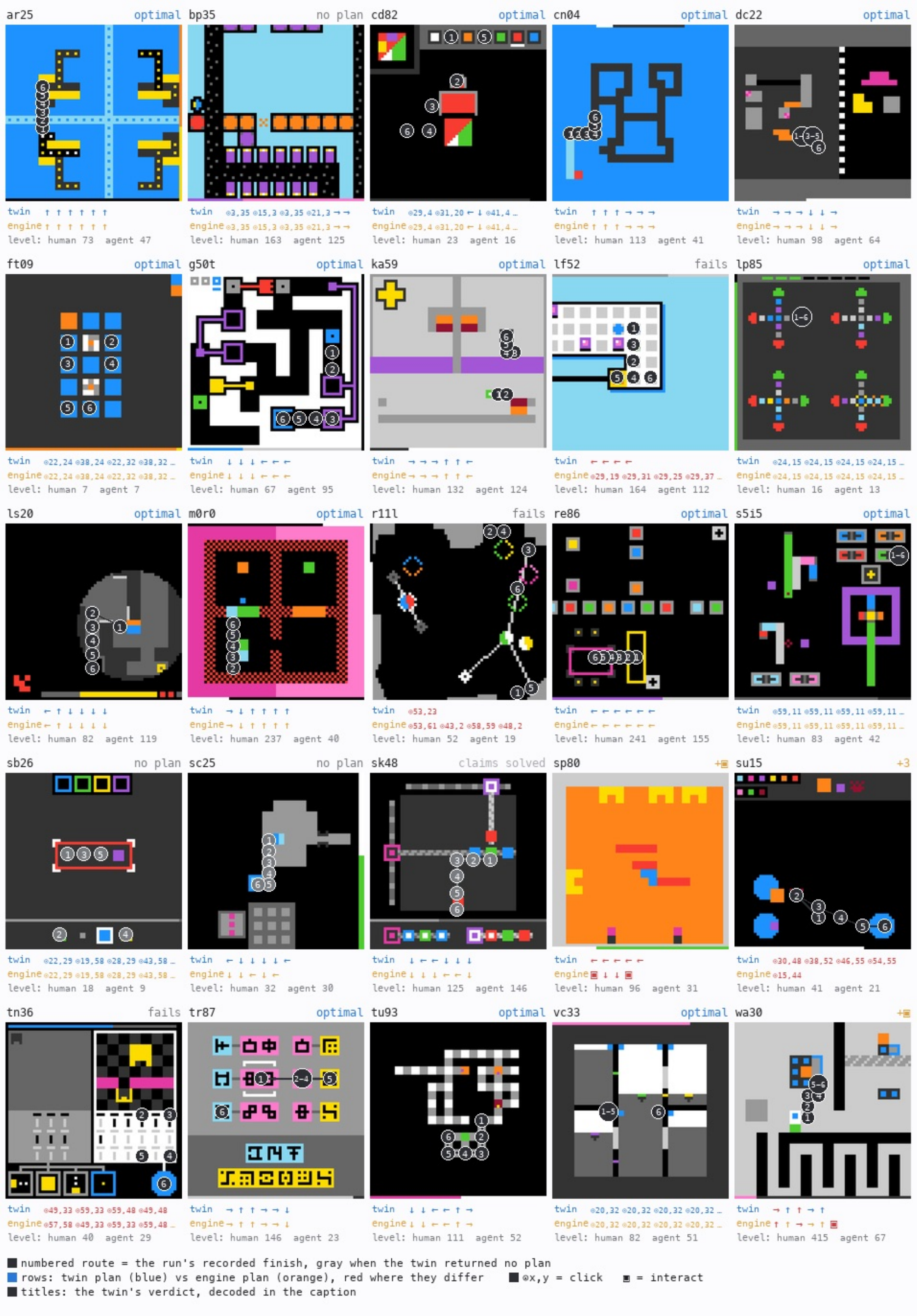}
\caption{The endgame gallery, one board per golden run (25), at the
moment a level is about to be won. Numbered dark circles: the run's
recorded finishing moves, never a plan. Gray circles and a gray twin
row mean the twin returned no plan; the row then repeats the recorded
route the circles trace. Stacked clicks share one badge;
bp35's route is suppressed, its camera re-anchoring every action. The
title is the twin's
verdict for that finish. \emph{optimal}: certified shortest. \emph{+3}:
three extra actions. \emph{fails}: failed in the engine. \emph{no plan}:
none returned. \emph{claims solved}: the twin returns the empty plan,
its goal firing early; never executed, so outside the 82. A plus sign
and a
filled square mark plans that stop one interact press short and win
once that press is appended. Rows below
each board: twin plan (blue) over engine plan (orange), red where they
differ; an ellipsis ends a row too wide for its panel.
The gray line counts the whole level's actions, the first-time human
baseline against the agent's run. tn36's engine row is the recorded
finish, its search defeated by the frame merging the text describes.
Across the 174 analyzed levels: 107 plans,
82 wins, all 82 certified, 80 exactly optimal.}
\label{fig:planchoice}
\end{figure*}

We investigate the forward dynamics of the world models \twin{} produces
against the dynamics of the real ARC-AGI-3 engine. We know from Appendix A
that each final twin recreates the recorded states one to one against the
true states the engine produced. This section asks how far the twin's
forward dynamics carry beyond that record. We built two experiments that
test the same bounded breadth-first search (BFS), once inside the final
twin and once inside the real engine, from the same restored states. The
search is the controlled constant: same algorithm, same candidate
moves, same depth. The candidate moves are the buttons and clicks the
run used, handed identically to both searches. Six actions out, both cases expand up to
$20{,}000$ states, the same cap the twin's planner used during the live
runs; from the start, both
expand up to $200{,}000$, with wall clock constraint. The
race grades whether \twin{} plans better than, equivalent to, or worse
than the ARC-AGI-3 engine's plans, over all 179 completed
levels. The first experiment, Figure~\ref{fig:paritymap}, runs the comparison
twice per level: once six recorded actions before the
level's win, inside the play-time depth-8 horizon, and once from the
level start, with the horizon expanded to that level's recorded
length plus two. The recorded finish, and any shorter one, therefore fits
inside the horizon; when a search misses, it ran out of budget, not of
depth. The
second experiment, Figure~\ref{fig:planchoice}, grades each run's
finishing plan on its real board against the engine's shortest path.

\subsection{The race over every completed level}

We test the two searches six actions out first
(Figure~\ref{fig:paritymap}, left block). The comparison covers 174 of the
179 completed levels. A level is compared only from a board proven
identical for both searches, rebuilt by replaying the run's recording
in a fresh engine and checking every frame. dc22's and tn36's replays
stop matching partway, at actions 358 of $1{,}218$ and 228 of
472. The cause is timing in the original recording, frames captured
while the board was still mid-change, not any version, seed, or
randomness difference, so their last five levels sit out as pale cells.
Over the 174 that remain, the engine finds a finish at 159, the
twin at 107, and both at 102.

Every disagreement in the block has a named cause. In the five
light-blue cells only the twin's search returns a plan, and every one
fails in the engine, a route the twin believes and the game
contradicts. On cn04's two levels the engine search hits its
$20{,}000$-node budget before any finish appears. On tn36's three the
engine search merges states that draw the same frame, and tn36 keeps
hidden state its frames do not show. The engine finds a finish the
twin misses on 57 levels, and the causes are the model failures
Appendix A already profiled: sc25's twin never sees its goal, wa30's
memorizes its history, and sk48's and bp35's searches exhaust on
boards their mechanics rewrite wholesale. On 10 levels, seven of
them sb26's, neither search succeeds within budget.

We then rerun the comparison from each level's start state under the
graduated horizon (right block). Depth becomes limiting: levels solved by
either search take a median of 17 recorded actions, whereas the 111 solved
by neither take 46; 35 engine searches also hit the wall-clock limit. From
the start, the engine finishes 61 levels and the twin 47, but 31 of the
twin's 32 wins are certified shortest. Two levels favor the twin outright:
re86 L1 and su15 L2 execute its 20- and 12-action plans exactly, while the
engine's search finds nothing because it merges identical frames. When hidden
state exceeds what the pixels show, the twin is the better search space. From
150 engine-generated detour states absent from its run, the twin returns
84 plans; 65 win in the engine, and 57 of 59 certified plans are
exactly optimal.



\FloatBarrier
\clearpage
\renewcommand{\topfraction}{0.98}
\renewcommand{\dbltopfraction}{0.92}
\renewcommand{\bottomfraction}{0.5}
\renewcommand{\textfraction}{0.02}
\renewcommand{\floatpagefraction}{0.75}
\renewcommand{\dblfloatpagefraction}{0.75}
\section{Appendix C: Token and Cost Accounting}

The benchmark scores completion and action efficiency; the compute
behind those scores is never tracked. Yet every scored action is bought
with unscored thinking. This appendix asks what one action really costs
in test-time compute, and what the world-model approach does to that
price. \twin{} pays 224k processed tokens per scored action, the
no-twin ablation 48k, and EWM 715k. Building the twin puts nearly
five times the thought behind each action and halves the actions the
harness takes. Against EWM that wins on both counts. Against the
ablation it is a trade: the twin does not save compute; it front-loads
the compute offline so the actions it submits online are well-thought and few.

\par\medskip\noindent\begin{minipage}{\columnwidth}
\centering
\includegraphics[width=\columnwidth]{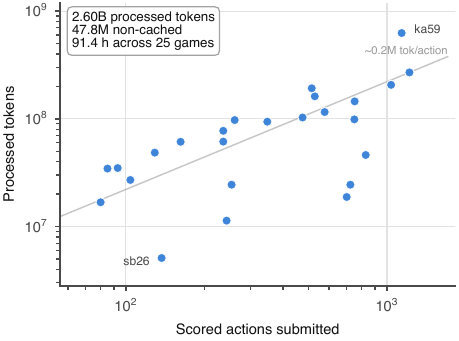}
\captionof{figure}{\twin{}'s action cost against its compute cost, one
point per game over all 25 runs. The horizontal axis shows scored
environment actions, which set the benchmark score; the vertical axis
shows processed test-time tokens, which the benchmark does not count.
Both axes are logarithmic. The diagonal marks the median rate, about
0.2M tokens per scored action. Games take between 80 and $1{,}219$
scored actions. Processed tokens run from 5.1M on sb26 to 625.3M on
ka59, the two labeled points. The summary box aggregates all 25 runs:
2.60B processed tokens, 47.8M of them non-cached, over 91.4 hours.
Efficiency in scored actions does not imply efficiency in test-time
compute.}
\label{fig:compute}
\end{minipage}\par\medskip

\begin{table}[t]
\centering\scriptsize
\renewcommand{\arraystretch}{0.92}
\setlength{\tabcolsep}{3.2pt}
\vspace*{-\abovecaptionskip}
\caption{Aggregate evaluation outcomes and compute. \twin{} and no-twin
ablation values are recomputed from their released run artifacts. EWM
values come from its published run. \emph{Tokens}: total processed
tokens. \emph{Actions}: scored environment actions. \emph{tok/act}:
processed tokens per scored action. Wall-clock time is omitted because
EWM does not publish it, and OPINE-World is absent because it reports no
token counts. \twin{} scores 93.3 on 2.60B total tokens against
EWM's 63.8 on 20.0B. The no-twin ablation runs cheaper, 48k tokens
per scored action and 1.06B in total; the twin's action efficiency is
bought with compute.}
\label{tab:computeagg}
\begin{tabular}{l r r r r r}
\toprule
System & Score & Levels & Tokens & Actions & tok/act \\
\midrule
\twin{} & 93.3 & 179/183 & 2.60B & 11,614 & 224k \\
No-twin ablation & 61.1 & 148/183 & 1.06B & 22,224 & 48k \\
EWM & 63.8 & 146/183 & 20.0B & 28,035 & 715k \\
\bottomrule
\end{tabular}

\medskip
\centering\scriptsize
\setlength{\tabcolsep}{4pt}
\captionof{table}{Per-game compute for \twin{}, all 25 games.
\emph{proc}: processed tokens, millions unless marked B. \emph{n-c}:
non-cached tokens, millions. \emph{h}: wall-clock hours. \emph{act}:
scored actions. \emph{tok/act}: processed tokens per scored action,
thousands. Wall clock spans 0.7 hours on sb26 to 17.8 on ka59, and
the thinking behind one action spans 34k tokens on wa30 to 549k on
ka59. The 25 runs total 2.60B processed tokens and 91.4 hours,
with non-cached tokens under 2\% of the total: the twin's thinking
runs almost entirely on cached context.}
\label{tab:computepergame}
\input{compute_pergame}
\end{table}

Compared with EWM, \twin{} improves the score by 29.5 percentage points on
$7.7\times$ fewer processed tokens (Table~\ref{tab:computeagg} and
Figure~\ref{fig:compute}). Compared with the no-twin ablation using an
off-the-shelf coding agent, \twin{} uses $2.4\times$ as many total tokens
and $4.7\times$ as many tokens per scored action, while improving the
score by 32 percentage points. Cached-context reads account for over 98\% of
processed tokens, with ka59 alone contributing 24\% of the total
(Table~\ref{tab:computepergame}).

\FloatBarrier
\clearpage
\section{Appendix D: Inside \textsc{Explore}}

\begin{figure*}[p]
\centering
\includegraphics[width=\textwidth]{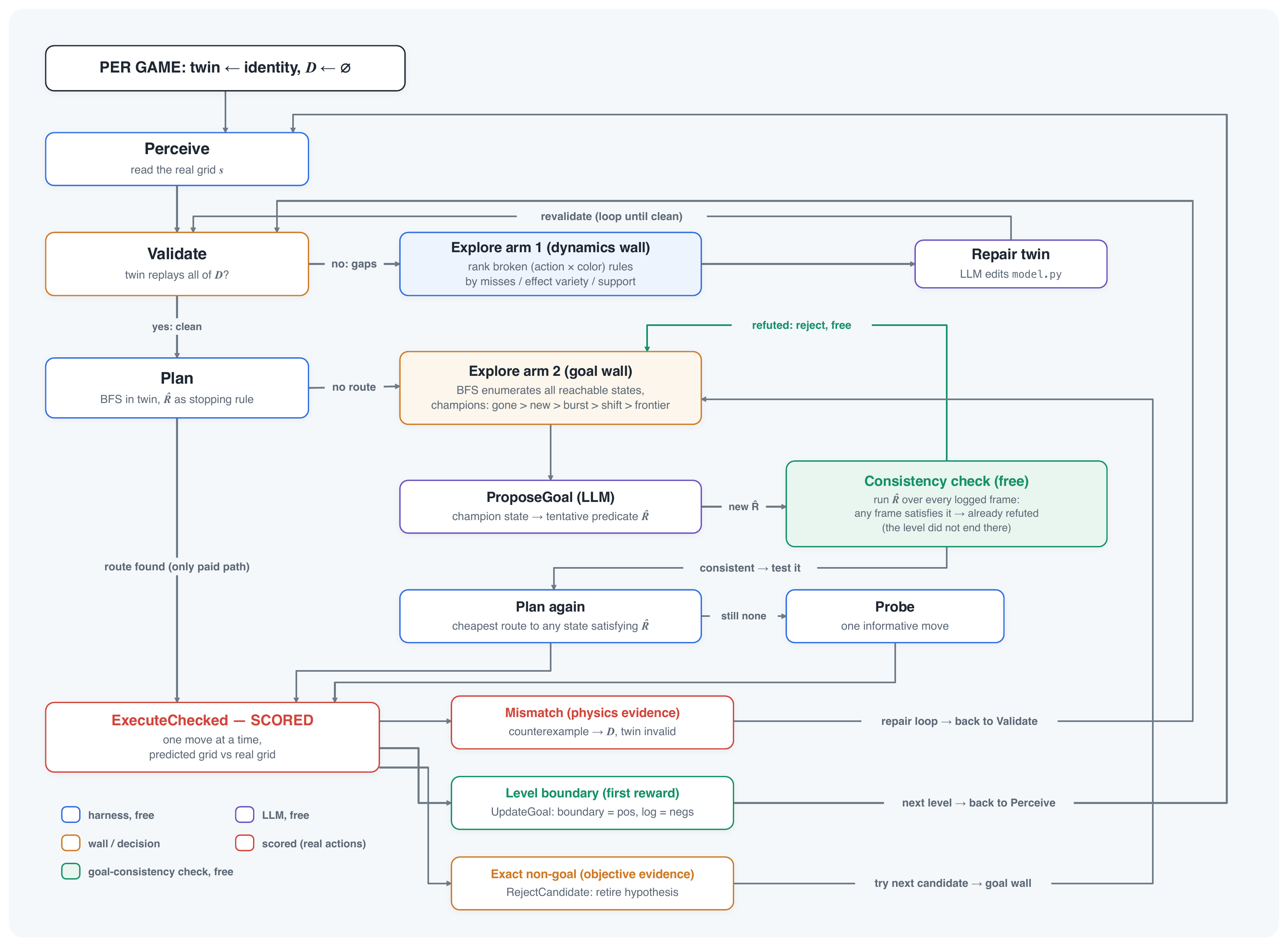}
\caption{End-to-end control flow for Algorithm~\ref{alg:twin}. Node
borders identify the responsible component: blue for the harness, violet for
the coding agent, amber for walls and decision points, red for scored
actions, and green for goal-consistency checks. The spine runs
\textsc{Perceive}, \textsc{Validate}, \textsc{Plan},
\textsc{ExecuteChecked}. The two explore arms are the cases of
Eq.~\ref{eq:explore}: arm 1 answers the dynamics wall with ranked
repairs, and arm 2 answers the goal wall with champion states, a
proposed predicate, and the consistency check before any plan is built
on it. Every node before the scored one is free, so a scored action is
taken only when nothing unscored remains. Three outcomes re-enter the
loop, each carrying evidence: a transition mismatch returns to
\textsc{Validate} with a counterexample, a level boundary returns to
\textsc{Perceive} with the first reward, and a predicted goal reached
without a boundary retires the candidate and restarts goal discovery.}
\label{fig:flow-overview}
\end{figure*}

\begin{figure*}[p]
\centering
\includegraphics[width=\textwidth]{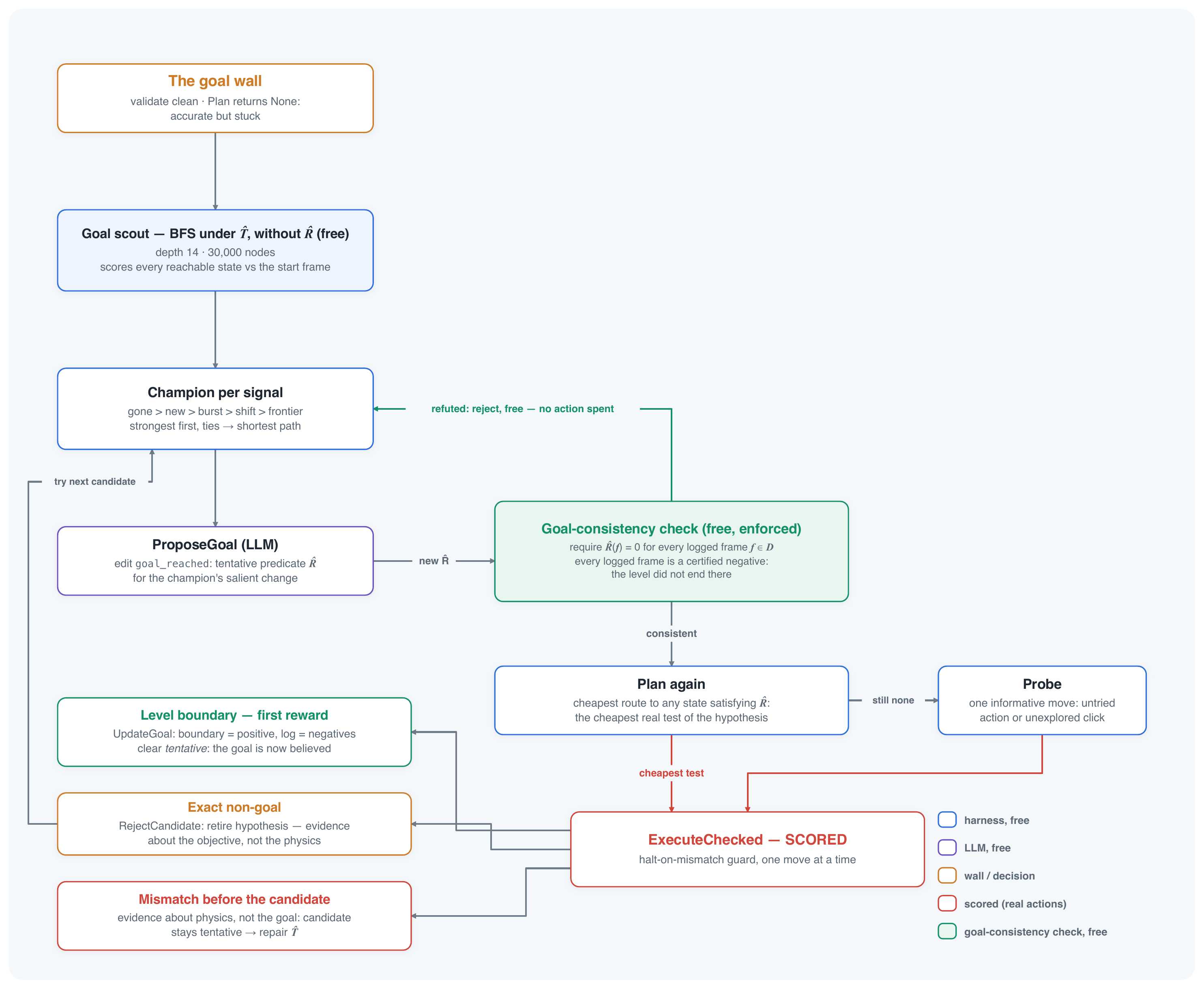}
\caption{Goal discovery and consistency checking, with node borders as in
Figure~\ref{fig:flow-overview}. At the goal wall the harness searches the
twin without a reward model $\hat R$, at depth 14 over $30{,}000$
nodes, and keeps the strongest state per progress signal, shorter paths
winning ties. The coding agent turns the top champion's salient change
into the tentative predicate $\hat R$. The green check requires $\hat R$
to reject every recorded frame, each one a certified negative, before any
plan is built on it. A survivor is tested along its cheapest route, one
scored action at a time under the halt-on-mismatch guard, and each
outcome settles a different question. A level boundary confirms the
candidate and becomes the first reward example. The predicted goal
reached without a boundary retires it: the physics were right and the
objective was wrong. A transition mismatch is the reverse: the physics
get repaired and the goal candidate stays tentative. The released harness enforces the green check as a hard
constraint, refusing any predicate that fails it; during the reported
runs the \textsc{UpdateGoal} prompt asked the agent to apply the same
check.}
\label{fig:flow-goal}
\end{figure*}

The Method named the two ways a run gets stuck, the dynamics wall when
the twin fails to replay the past, the goal wall when it replays
everything yet reaches no goal. \textsc{Explore} answers the live wall
with one unscored
diagnostic query, and one case split chooses it. Let
$c=\textsc{Validate}(\hat T,\mathcal{D})$ be the counterexamples the
current twin leaves when it replays the recorded history
$\mathcal D$. Then
\begin{equation}
\textsc{Explore} =
\begin{cases}
\text{dynamics gaps} & c\neq\varnothing,\\[1pt]
\text{goal candidates} & c=\varnothing,\ \text{no plan},\\[1pt]
\varnothing & \text{plan found.}
\end{cases}
\label{eq:explore}
\end{equation}
A nonempty $c$ means the dynamics are wrong, so the query targets the
worst gap: the action and object pair the twin mispredicts most. An empty $c$ with no plan means the twin is right
and the goal is missing, so the query proposes a goal candidate. A found
plan voids any exploration. Figure~\ref{fig:flow-overview} draws
the complete control loop, and Figure~\ref{fig:flow-goal} expands its
goal-discovery branch.

\paragraph{Dynamics diagnostics.} The dynamics branch, the repair loop of
Figure~\ref{fig:flow-overview}, replays the recorded
interaction history $\mathcal{D}$ once and groups every changed cell by the
action taken and by the colors it held before and after the change. Clicks
aggregate across their coordinates, background cells are excluded, and what
remains is one context per action on one color of object. Three statistics
rank the contexts, each tied to a remedy. Misprediction rate dominates: a
rule the twin gets wrong is repaired first. Effect variety, the number of
distinct effect signatures, flags outcomes that vary under one rule and
calls for a split on a before-state feature. Support count marks a rule
seen only once, worth one probe to confirm it generalizes. The hints follow
the self-debugging pattern \citep{chen2024selfdebug} inside a
counterexample-guided loop in the CEGIS line
\citep{solarlezama2008sketching}. The grouping operates over object-level
abstractions rather than game-specific rules, so nothing in it is
hard-coded to ARC-AGI-3.

\paragraph{Goal discovery.} When the twin validates and no plan reaches a
goal, the branch Figure~\ref{fig:flow-goal} expands, the planner's
breadth-first search runs again inside the twin with a
different job: instead of stopping at the first goal state, it scores
every reachable state against the starting grid on five signals, ranked
in this order:
\begin{itemize}\itemsep1pt
\item \texttt{color\_gone}: a start color vanishes entirely, the shape of
 something consumed or completed.
\item \texttt{color\_new}: a color absent from the start appears, the
 shape of an unlock or a reveal.
\item \texttt{local\_burst}: $\ge 5$ cells change inside one compact
 bounding box while the rest stays static, a localized event.
\item \texttt{big\_change}: over a quarter of all cells change at once, a
 scene shift.
\item \texttt{frontier}: the fallback, the most-different reachable state.
\end{itemize}
Scoring everything needs more room than finding one route, so the budget
grows from the planner's depth 8 and $20{,}000$ states to depth 14
and $30{,}000$. The search keeps the cheapest strong example of each
signal, so the candidates come out diverse and cheapest first.
These fixed signals are visual-change heuristics used only to rank candidate
states; they do not encode game-specific objects, actions, or goals. The
coding agent must still infer a goal predicate from the selected state and
validate it against the recorded interaction history.
A level boundary encountered along the cheapest route to a top-scoring state
provides the first positive example for \texttt{goal\_reached}. Goal discovery
is a Go-Explore-style novelty search \citep{ecoffet2019goexplore} serving
WorldCoder's optimism constraint \citep{tang2024worldcoder}: propose a reward
believed to be reachable, then use it to direct exploration.

\paragraph{Goal-consistency check.} One invariant, the green check in
Figures~\ref{fig:flow-overview} and~\ref{fig:flow-goal}, filters every
candidate predicate: \texttt{goal\_reached} must return false on every
frame in the recorded history $\mathcal{D}$. The invariant is sound because ARC-AGI-3
replaces the winning frame when reporting completion, so a true goal frame
never enters the record and every recorded frame is a certified negative.
A candidate that fires on any of them is rejected before a plan is built
on it. The check instantiates WorldCoder's consistency condition for
proposed rewards \citep{tang2024worldcoder}.

\paragraph{Submission check.} Only \texttt{submit.py} issues a scored
action, the red node of Figure~\ref{fig:flow-overview}, and it demands
three things first: a rationale tag, a twin that
validates against the complete recorded history, and a next-frame
prediction committed before the outcome is observed. The tag declares
what the action is for:
\begin{equation}
\text{scored action} =
\begin{cases}
\texttt{MODEL:} & \text{execute the twin's plan},\\[1pt]
\texttt{PROBE:} & \text{test an uncertain rule.}
\end{cases}
\label{eq:submit}
\end{equation}
The tag is the agent's own reasoning, declared at submission rather than
checked by the harness, which enforces only the validation and the
committed prediction. The declaration puts the agent's certainty on
record: \texttt{MODEL:} says the twin is trusted enough to act on,
\texttt{PROBE:} says information about the environment is still missing.
The submission log keeps that split, so every scored action is auditable after
the fact as either exploitation or evidence-buying.

\FloatBarrier
\section{Appendix E: Every Scored Action}

\begin{figure}[H]
\centering
\includegraphics[width=\columnwidth]{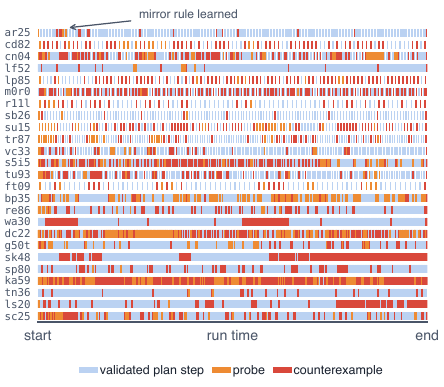}
\caption{Every scored action in the 25 runs, one row per game,
with games sorted by score. The horizontal axis is run time, measured in
scored actions and normalized so that every run spans the same width. Blue
marks a validated plan step whose prediction was correct, amber a deliberate
probe, and red a plan step whose failed prediction produced a counterexample.
The arrow marks the action at which ar25 learned its mirror rule. Probes
account for 7.1\% of the $11{,}562$ actions the agent chose, and the
remaining 92.9\% execute plans tested in the twin.}
\label{fig:actions}
\end{figure}

Figure~\ref{fig:actions} turns Eq.~\ref{eq:submit}'s tag into a run-level
view: every scored action of the 25 runs, one row per game. Validated
plan steps dominate the rows. Of the probes, nineteen in twenty
distinguish between competing dynamics hypotheses; only one in twenty
searches for the goal directly. Exploration is front-loaded, 19.1\% of
first-level actions and 5.3\% of second-level, and probes reappear in
deeper levels when new mechanics create fresh uncertainty.

The rows also separate learning regimes. After ar25 learns its mirror
rule, 223 of its remaining 235 actions are validated plan steps. ka59
and sk48 never settle, interleaving successful plans with counterexamples
to the end. \twin{} does not wait for a globally accurate model: the
halt-on-mismatch guard runs each plan until its first wrong prediction,
and that prediction becomes the next repair.

\section{Appendix F: One Base Model, From Direct Play to \twin{}}

Played directly, the base model GPT 5.6 Sol scores 7.8 on the
leaderboard. The nearest alternative to a world model is better memory,
and in July 2026 OpenAI reported two memory settings for this same model:
retaining its private reasoning across moves, and compacting older
history \citep{openai2026twosettings}. On OpenAI's own evaluation, where
direct play starts higher at 13.3, the two settings together reach
38.3 on roughly six times fewer output tokens. Both settings keep the
model's working state alive across moves. Prime Agent
\citep{primeintellect2026primeagent} pushes that principle furthest. It holds
the interaction history as a variable inside a persistent interpreter, and it
lets the agent rewrite its own prompt, skills, and sub-agents mid-run. Prime
Agent reaches 78.3, the strongest score reported on this base model without a
world model, ahead of the off-the-shelf Codex harness at 61.1. \twin{}
externalizes the same working state as executable code, which survives every
agent restart and is validated against the recorded interaction history.
Managing that state carries the base model from 7.8 to 78.3, and the final
15.0 points to \twin's 93.3 are what a validated world model adds
(Table~\ref{tab:directplay}).

\begin{table}[H]
\centering\footnotesize
\setlength{\tabcolsep}{4pt}
\caption{Direct play and world-model methods on the 25-game public set,
on the benchmark's action-efficiency score. Row one is the leaderboard's
direct-play score. Rows two and three are OpenAI's own runs as published
\citep{openai2026twosettings}: row two the official evaluation, row three
the same run plus the two memory settings, retained reasoning and history
compaction. The last three rows are from Table~\ref{tab:agg} and hold the base
model fixed while changing only the harness. The three
direct-play scores are as published, not rescored here. The strongest of
them remains 55.0 percentage points below \twin's 93.3.}
\label{tab:directplay}
\begin{tabular}{l r}
\toprule
Configuration (same base model) & Score \\
\midrule
Direct play, leaderboard & 7.8 \\
Direct play, official evaluation, OpenAI's run & 13.3 \\
\quad + retained reasoning + compaction & 38.3 \\
Off-the-shelf Codex harness & 61.1 \\
Prime Agent harness & 78.3 \\
\twin{} & \textbf{93.3} \\
\bottomrule
\end{tabular}
\end{table}

\section{Appendix G: The Learned Twins}

Each run leaves its world model behind as code, a \texttt{model.py} the
agent wrote and optumized. What the twin learned is not hidden in
weights: every rule it believes, and the reasoning behind each repair,
reads as source and docstring.
Listing~\ref{lst:ka59} (Appendix~K) shows both halves of that learning in ka59, rule
induction and repair from a counterexample. The docstring, the agent's
own words, states a conservative discipline: effects stay no-ops until
they are observed, and every new rule must keep reproducing every
transition in the recorded interaction history. From a few observations
the twin recovers the action semantics exactly, each arrow moving the
token by three pixels. A later counterexample exposes a perception edge
case. With the token in the corner pocket of a plus, only three rim cells
stay visible, so the repair lowers the detector's threshold from four
cells to three.

Listing~\ref{lst:repair} (Appendix~K) shows the same behavior in ft09, where learning
starts at the first possible moment. The very first scored action returns
a counterexample, two wrong cells in the bottom action-budget bar, a
mechanic the twin had not modeled. Three actions later level 0 ends, the
unsolved level-1 board appears, and the goal predicate fires on it
anyway. What the predicate had memorized was a picture of one finished
board, not the goal. The agent rewrites it as a constraint, and the same
board now fails it.

\begin{figure*}[t]
\centering
\includegraphics[width=\textwidth]{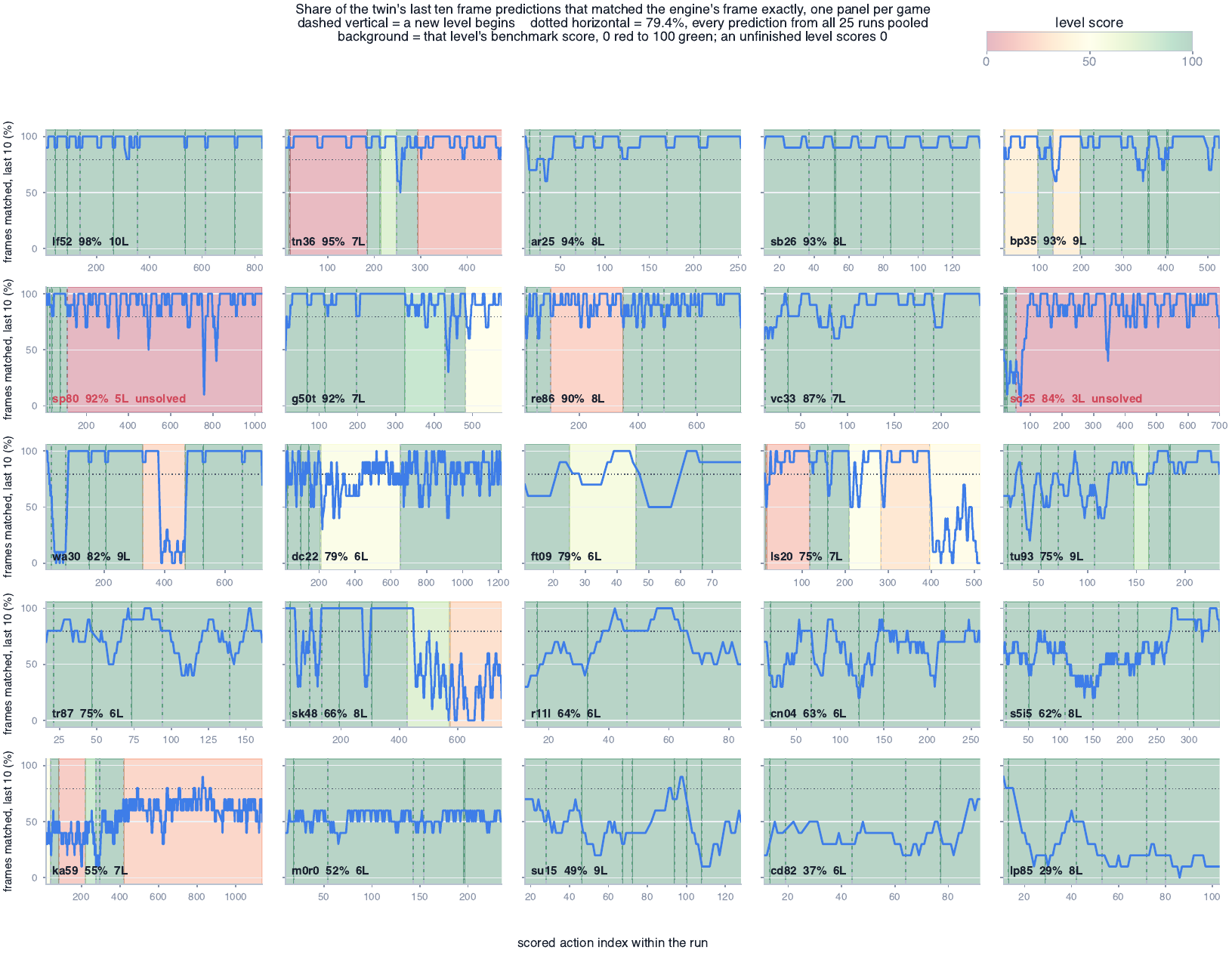}
\caption{Each panel displays a single ARC-AGI-3 game, where the twin
predicts the next frame and compares that prediction against the
ARC-AGI-3 engine's frame given the same action. Each data point is a
rolling average, the ratio of exactly matched frames over the ten most
recent predictions, so one miss moves the curve by ten percentage
points. The first data point on the line is produced after ten actions
have been submitted to the simulator, when the first ratio is
recorded. The x axis is the action index for that game, so the value
at 100 shows the frame prediction accuracy at the 100th action.
Every prediction is hash-stamped before its action is sent, committed
ahead of the answer. Vertical dashed lines are level boundaries. The
background is a heat map of each level's benchmark score, red at 0
to green at 100 on the printed scale; an unfinished level scores
0, so the two unsolved runs end in deep red. The dotted line pools
all 25 runs: $8{,}532$ of $10{,}744$ predictions with a recorded
outcome were exact, 79.4\%. Each title gives the run's exact rate over its whole history, the
share of all its predictions, first to last, that matched exactly,
plus levels cleared, red for the two unfinished games. The
title counts every prediction while the curve counts the last ten, so
ls20 reads 75\% even as its final window falls. Read each curve as
that game's training signal. Flat at the top on green, lf52 and ar25,
means the rules were learned once and the game came easy. Dips at a
boundary that recover, sb26 and g50t, are new mechanics becoming
repairs. Red under a flat curve, tn36, is accuracy spent hunting the
goal rather than the rules. A curve that keeps dropping and rising,
ka59's long middle or sc25's opening, means the rules keep breaking
and the game is hard; sc25 and sp80 stay unsolved.}
\label{fig:learning2}
\end{figure*}

Figure~\ref{fig:learning2} scales the mispredict-then-repair cycle of
Listings~\ref{lst:ka59} and~\ref{lst:repair} to all 25 games. Before every \texttt{MODEL}-tagged
action the twin predicts the full next frame, and the harness stamps the
prediction's hash before the action is sent, so the claim cannot be
revised. A prediction is exact when the twin's predicted frame and the
frame the engine returns agree on all $4{,}096$ cells. Each run draws
one curve, a running average over its last ten predictions: the fraction
that were exact. A single miss moves it by ten percentage points, so no error
hides.
Most curves start rough, because a fresh twin knows nothing, and climb
as each miss becomes a repair. Dips arrive with new levels, where fresh
mechanics make the current rules wrong again. The spread is wide, 29\%
on lp85 to 98\% on lf52 over whole runs, and both games were cleared.
Low accuracy does not doom a run because errors are contained. A plan
halts at its first wrong prediction, so one bad rule wastes at most one
action and hands back a counterexample. ls20 is the extreme, its final
window collapsing while the run still reads 75\% and finishes. The
curves never flatten into a solved model; prediction and repair never
stop.

Putting each curve beside its background scores, the figure separates
good training signals from bad ones. The modal good signal is the one
the eye expects, flat and high on green. lf52, ar25, sb26, re86, and
bp35 all hold above 90\% and win every level at or near human pace,
each miss repaired the moment it appears. The rule fails as
arithmetic, though. Across the 25 runs, the share of a curve's
windows at 90\% or higher has no correlation with the game's score,
because the same flat-high shape sits on the two worst panels. tn36
holds 95\% and scores 69.7. sp80 holds 92\% and never finishes.
A high curve is a good signal only when its tint agrees. The opposite
corner teaches the same lesson from below. cd82, lp85, and m0r0 run at
37\%, 29\%, and 52\% and still score 100, because a miss is
cheap by construction: the plan halts, one action is spent, and the
miss returns as a counterexample. What a bad signal actually looks
like is neither height nor noise but failure to convert. ka59's dips
never recover, its curve spending a median of 241 actions below
80\% after each new level, repairs that do not stick. sk48's tail
collapses and stays down, its last levels sliding red. The subtlest
bad signal is the healthiest looking one: sp80 and sc25 end on long
red stretches under high curves, the model fine, the goal missing.
Height says how much the twin knows. Recovery says how fast it
learns. Only the tint says whether what the twin learned turned into
benchmark score, and the benchmark counts nothing else.

\begin{figure*}[p]
\centering
\includegraphics[width=\textwidth]{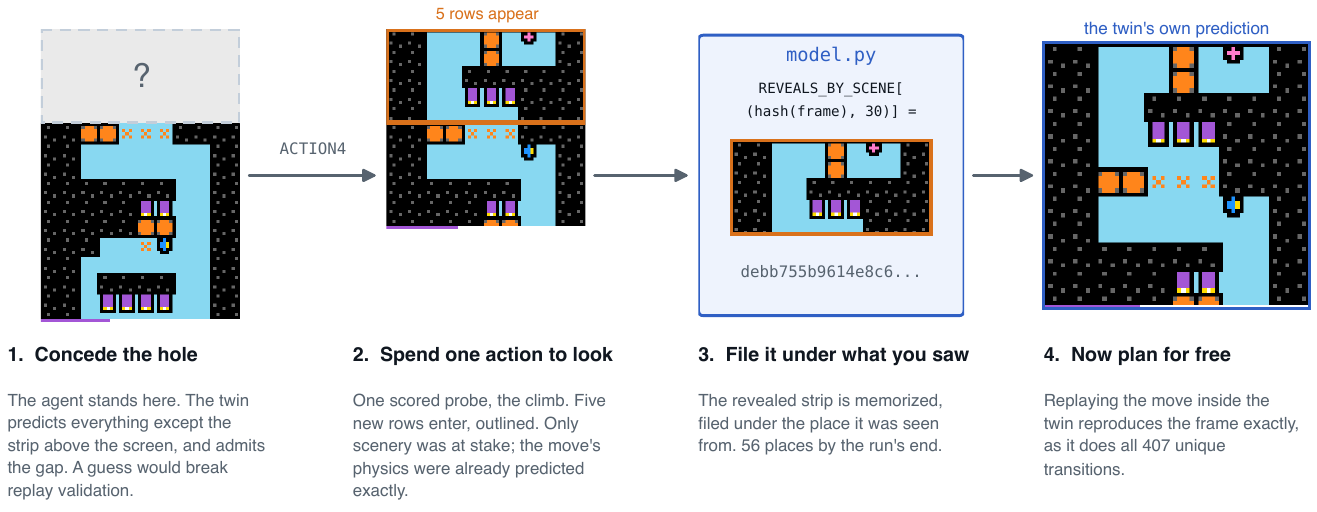}
\caption{bp35's answer to partial observability, pay once per new
place. Left to right: the twin predicts everything except the strip
about to appear and concedes it. One scored probe looks. The revealed
strip is filed in \texttt{model.py} under a hash of the frame it was
seen from, 56 entries by the run's end. The planner then searches
through that camera move at no cost. Rebuilt this way, the twin
reproduces all 407 unique recorded transitions exactly.}
\label{fig:bp35loop}
\vspace{16pt}
\includegraphics[width=0.78\textwidth]{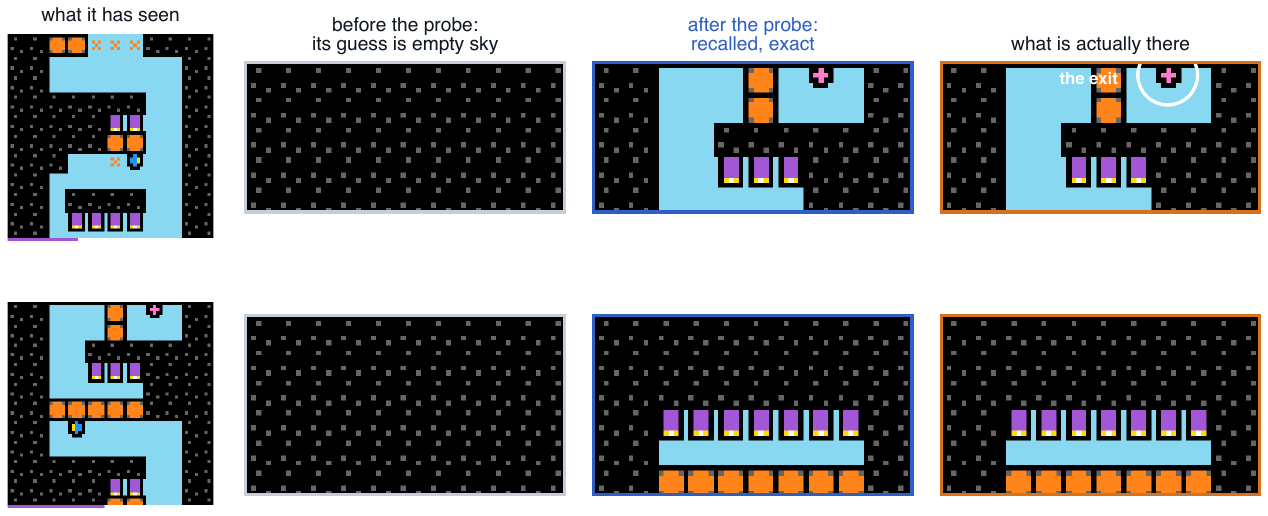}
\captionof{figure}{What the twin expects above the seen windows, and
what one probe buys, two camera moves from the run. Each row reads
left to right: the window the agent stands in, the guess with the
memorized strip erased, the prediction once the strip is filed, and
the truth. Before the probe the guess is always empty sky. After it
the prediction is exact in both rows. In the top row the
hidden strip holds the exit itself, ringed, so no plan can aim at the
goal until the probe buys the truth. Pooled over the 54 camera moves
that needed the cache, the sky guesses land 55\% of pixels but
0.1\% of the geometry, and 50 of the 54 hold not one platform,
hazard, or exit. The map is recall, not inference. The twin replays
any strip it has paid to see and expects nothing but sky where it has
not looked. One probe per place is the price.}
\label{fig:bp35prior}
\end{figure*}

\paragraph{A game the twin can never fully see.} \twin{} handles
fully observed deterministic games, and bp35 is the evidence it
stretches to a class of partially observed ones: the screen shows
10.7 rows of a 28-row map, the camera scrolls with the player, and
the exit sits 16 rows above the opening frame, in space the agent
has no pixel of. The run clears 9 of 9 levels in 529 actions
against the 651-action human baseline, seeing about a third of the
map at a time. The reason it works is that bp35 hides map, not state. The camera is a pure function of the
player's row and nothing latent moves on its own, so the unseen part
of the level is a static piece of the transition function, learnable
once and then checkable forever. The twin learns it by paying once per
new place (Figure~\ref{fig:bp35loop}). It predicts the whole next
frame except the strip about to scroll into view, and concedes that
strip. The agent spends one scored action to look. The revealed strip
is pasted into \texttt{model.py} under a hash of the frame it was seen
from, 56 entries by the run's end, and that place never costs again.
Most of the twin is still rules rather than recall. Deleting all 56
strips breaks only 54 of the 407 unique recorded transitions;
learned physics carries the other 353. The weak point is the prior
over unseen space (Figure~\ref{fig:bp35prior}). Asked what sits above
the seen windows, the twin answers empty sky.

\clearpage
\onecolumn
\begingroup
\centering\small
\setlength{\tabcolsep}{7pt}
\renewcommand{\arraystretch}{1.22}
\captionof{table}{Per-game results on all 25 public games. act = scored actions. lvl =
levels cleared\,/\,total. score = benchmark action-efficiency score, with all systems
evaluated against the same human baselines. \twin{}'s unfinished games are shown as-is (sc25 $3/6$, sp80 $5/6$).
\textbf{Bold} marks \twin's closing results, the games won and the mean score. }
\label{tab:main}
\begin{tabular}{l r r c c r c c r c c r c c}
\toprule
& Human & \multicolumn{3}{c}{\textbf{\twin{} (ours)}} & \multicolumn{3}{c}{OPINE-World} & \multicolumn{3}{c}{EWM} & \multicolumn{3}{c}{Codex} \\
\cmidrule(lr){3-5}\cmidrule(lr){6-8}\cmidrule(lr){9-11}\cmidrule(lr){12-14}
Game & act & act & lvl & score & act & lvl & score & act & lvl & score & act & lvl & score \\
\midrule
ar25 & 748 & 254 & 8/8 & 100.0 & 381 & 8/8 & 100.0 & 264 & 8/8 & 100.0 & 299 & 8/8 & 100.0 \\
bp35 & 651 & 529 & 9/9 & 100.0 & 512 & 2/9 & 2.6 & 342 & 4/9 & 17.3 & 1,303 & 5/9 & 2.5 \\
cd82 & 171 & 93 & 6/6 & 100.0 & 161 & 6/6 & 100.0 & 130 & 6/6 & 100.0 & 150 & 6/6 & 98.3 \\
cn04 & 789 & 261 & 6/6 & 100.0 & 263 & 6/6 & 100.0 & 448 & 6/6 & 100.0 & 237 & 6/6 & 100.0 \\
dc22 & 1,228 & 1,219 & 6/6 & 97.1 & 1,479 & 6/6 & 82.6 & 1,842 & 4/6 & 36.4 & 2,457 & 5/6 & 55.5 \\
ft09 & 208 & 80 & 6/6 & 100.0 & 111 & 6/6 & 100.0 & 157 & 6/6 & 95.1 & 183 & 4/6 & 54.8 \\
g50t & 879 & 577 & 7/7 & 90.2 & 757 & 7/7 & 61.7 & 554 & 7/7 & 100.0 & 1,759 & 5/7 & 33.7 \\
ka59 & 730 & 1,138 & 7/7 & 73.3 & 1,076 & 6/7 & 62.5 & 1,099 & 7/7 & 63.0 & 1,086 & 7/7 & 52.9 \\
lf52 & 1,339 & 829 & 10/10 & 100.0 & 593 & 3/10 & 4.2 & 2,646 & 6/10 & 34.9 & 2,679 & 4/10 & 12.4 \\
lp85 & 388 & 104 & 8/8 & 100.0 & 110 & 8/8 & 100.0 & 226 & 8/8 & 100.0 & 574 & 8/8 & 82.0 \\
ls20 & 776 & 515 & 7/7 & 100.0 & 959 & 7/7 & 71.2 & 714 & 7/7 & 98.3 & 977 & 3/7 & 9.9 \\
m0r0 & 1,107 & 236 & 6/6 & 100.0 & 259 & 6/6 & 100.0 & 2,573 & 5/6 & 71.6 & 311 & 6/6 & 100.0 \\
r11l & 233 & 85 & 6/6 & 100.0 & 128 & 6/6 & 100.0 & 227 & 6/6 & 74.3 & 467 & 5/6 & 58.4 \\
re86 & 1,255 & 750 & 8/8 & 100.0 & 850 & 8/8 & 100.0 & 1,754 & 4/8 & 26.4 & 865 & 8/8 & 99.9 \\
s5i5 & 638 & 348 & 8/8 & 100.0 & 638 & 4/8 & 27.3 & 3,028 & 3/8 & 0.7 & 361 & 8/8 & 100.0 \\
sb26 & 213 & 137 & 8/8 & 100.0 & 214 & 8/8 & 89.8 & 150 & 8/8 & 100.0 & 214 & 8/8 & 86.8 \\
sc25 & 350 & 701 & 3/6 & 32.7 & 256 & 6/6 & 84.0 & 1,107 & 3/6 & 20.1 & 701 & 5/6 & 44.8 \\
sk48 & 1,070 & 752 & 8/8 & 87.2 & 596 & 4/8 & 21.3 & 2,823 & 5/8 & 30.2 & 2,141 & 3/8 & 7.3 \\
sp80 & 518 & 1,037 & 5/6 & 82.1 & 369 & 6/6 & 100.0 & 786 & 1/6 & 5.5 & 1,037 & 2/6 & 1.5 \\
su15 & 361 & 129 & 9/9 & 100.0 & 334 & 9/9 & 92.1 & 284 & 9/9 & 88.7 & 723 & 6/9 & 37.7 \\
tn36 & 317 & 475 & 7/7 & 69.7 & 417 & 7/7 & 68.4 & 3,058 & 6/7 & 26.2 & 635 & 5/7 & 26.0 \\
tr87 & 414 & 162 & 6/6 & 100.0 & 212 & 6/6 & 100.0 & 540 & 6/6 & 86.3 & 494 & 6/6 & 88.1 \\
tu93 & 462 & 236 & 9/9 & 100.0 & 272 & 9/9 & 100.0 & 192 & 9/9 & 100.0 & 264 & 9/9 & 98.4 \\
vc33 & 447 & 243 & 7/7 & 100.0 & 427 & 7/7 & 92.4 & 1,596 & 3/7 & 19.4 & 199 & 7/7 & 100.0 \\
wa30 & 1,843 & 724 & 9/9 & 100.0 & 1,465 & 9/9 & 100.0 & 1,495 & 9/9 & 100.0 & 3,101 & 9/9 & 77.2 \\
\midrule
Won\,/\,mean & & & \textbf{23} & \textbf{93.3} & & 20 & 78.4 & & 14 & 63.8 & & 13 & 61.1 \\
\bottomrule
\end{tabular}
\par
\endgroup

\clearpage
\section{Appendix H: Action Counts by Level}

Tables~\ref{tab:perlevela} and~\ref{tab:perlevelb} are the raw ledger
behind every score in the paper: the actions each system spent on each
level, for the human reference, EWM, OPINE-World, and \twin{}. Columns
L1 through L10 follow the level order within each game. A blank cell
means the game has no such level or the row's system never reached it.
For an unfinished run, the first uncleared level reports the actions
spent there before the run ended. Human, EWM, and OPINE-World counts are
their published values, and \twin{} counts come from our recorded action
histories. The bolding gives the ledger's summary: on 122 of the 169 level
cells where at least one rival system also reached the level, \twin{} spent the
fewest actions or tied for fewest, 92 of them outright.

\begingroup
\centering\scriptsize
\setlength{\tabcolsep}{13pt}
\renewcommand{\arraystretch}{0.72}
\captionof{table}{Action counts by level for the first thirteen games, ordered
as in the OPINE-World appendix. \textbf{Bold} marks the fewest actions for that
level among EWM, OPINE-World, and \twin{}, ties included. The human row is the
reference and competes for no bold.}
\label{tab:perlevela}
\input{perlevel_a}
\par
\endgroup

\clearpage
\begingroup
\centering\scriptsize
\setlength{\tabcolsep}{13pt}
\renewcommand{\arraystretch}{0.72}
\captionof{table}{Action counts by level for the remaining twelve games,
continuing Table~\ref{tab:perlevela} with the same columns and cell
conventions. \twin{} spent 647 actions on sc25 L4 and 928 on sp80 L6
before those runs ended. Bold follows the same rule as Table~\ref{tab:perlevela}.}
\label{tab:perlevelb}
\input{perlevel_b}
\par
\endgroup

\section{Appendix I: Reproducibility Details}

\begin{figure}[ht]
\centering
\includegraphics[width=0.94\textwidth]{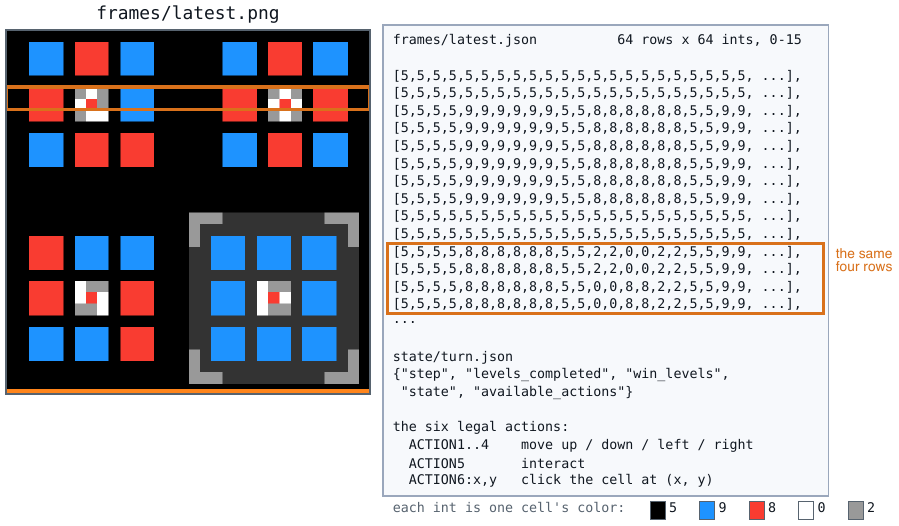}
\caption{The observation, as the agent receives it. Each step the
harness writes the current board to \texttt{frames/latest.json}, 64
rows of 64 integers 0 to 15, with a rendered
\texttt{frames/latest.png} beside it and a \texttt{state/turn.json}
carrying step, level, and the legal actions. Shown is ft09's first
frame from the golden run, rows and columns truncated for width. The
agent's observation is the integers, read as text. The rendered image
is written beside them for the record; the agent has no vision
channel, though a few runs parsed the image's pixels with code.}
\label{fig:observation}
\end{figure}

\noindent
\begin{minipage}[t]{0.48\textwidth}
\vspace{0pt}\normalsize
Table~\ref{tab:repro-config} reports the evaluation settings. The environments
and harness are deterministic. \texttt{gpt-5.6-sol} is accessed through a
hosted service without an exposed random seed, so its exact token sequences
are not reproducible. Reproducibility rests instead on the audit trail: one
scored run per game, a hash-committed prediction before every action the agent
submitted, and complete action histories. Every artifact is public. The
project site, \url{https://arc-agi-3-twin.vercel.app/}, replays all 25
runs action by action. The source and run data live at
\url{https://github.com/Alexyskoutnev/TWIN-ARC-AGI-3}.
Figure~\ref{fig:observation} shows what the agent actually perceives
each step: no rendered pixels, a page of integers.
\end{minipage}\hfill
\begin{minipage}[t]{0.48\textwidth}
\vspace{0pt}
\begingroup
\captionsetup{type=table,justification=raggedright,singlelinecheck=false}
\captionof{table}{Evaluation configuration. Every setting was fixed before
the first run and held constant across the 25 games.}
\label{tab:repro-config}
\centering\footnotesize
\setlength{\tabcolsep}{4pt}
\renewcommand{\arraystretch}{0.96}
\begin{tabular}{@{}>{\raggedright\arraybackslash}p{0.34\linewidth}
                    >{\raggedright\arraybackslash}p{0.58\linewidth}@{}}
\toprule
Setting & Value \\
\midrule
Planning search & depth 8, $20{,}000$ nodes \\
Goal discovery search & depth 14, $30{,}000$ nodes \\
Scored action budget & $2\times$ the game's human baseline \\
Agent turn timeout & 900 s \\
Agent relaunch cap & 50 \\
Reasoning effort & maximum \\
Parameter search & none \\
\midrule
Workstation & Apple M5 Max, 64~GB, macOS~26.5 \\
Software & Python 3.12, Codex CLI 0.144.6 \\
Hosted model & \texttt{gpt-5.6-sol} \\
\bottomrule
\end{tabular}
\endgroup
\end{minipage}

\clearpage
\twocolumn
\normalsize
\section{Appendix J: What the Twins Believe Winning Looks Like}

The \twin{} harness plans by searching a twin until its goal predicate accepts
a state. That predicate carries the plan's whole purpose. Search stops at the
first state it accepts, so every committed plan is a bet that its final state
ends a level. When the bet is wrong, the harness spends the plan's scored
actions and no level arrives. The main text grades the predicate once per
level, reporting the first committed goal hypothesis correct on 156 of
179 completed levels (87.2\%). One verdict per level leaves the stopping
behavior unmeasured, because breadth-first search tests the predicate at
every state it expands. We therefore grade every goal claim the twins
committed. A claim is correct when the action carrying it completes a level.
Over the $11{,}557$ graded actions the predicate reaches $\frac{138}{179} = 0.771$ recall at $\frac{138}{646} = 0.214$ precision, where 138 counts the wins the predicate accepted, 179 every completed level, and 646 every state it accepted. It finds most wins and also accepts many states that complete nothing. That looseness is the paper's residue measured at its finest grain, once per action instead of once per level. The twins learn how the world moves more reliably than they learn what winning in it means.

To grade the claims the twins made while playing, we read the submission
log, the entry the harness writes before each scored action. Each entry
records
\texttt{goal\_reached} applied to the twin's predicted next frame, committed
before the outcome arrives, so revision after the fact is impossible. Level
boundaries come from the recorded per-level action counts, and the
engine-verified replays behind Appendix~B confirm 150 of the 179
boundaries independently.

\begin{table*}[!t]
\centering\small
\setlength{\tabcolsep}{9pt}
\renewcommand{\arraystretch}{1.05}
\captionof{table}{Every goal claim across the 25 runs, graded by what
the action did. A goal claim is mechanical, not conversational. At each
submission the harness runs the twin's \texttt{step} on the current
frame and the action about to be sent, giving the twin's predicted next
frame, the same prediction Figure~\ref{fig:learning2} grades. It then
runs the agent-written \texttt{goal\_reached} on that predicted frame.
If \texttt{goal\_reached} returns true, the twin is claiming that this
action ends the level. The claim is logged before the outcome arrives,
so it cannot be revised. Every action then lands in one of four cells.
If \texttt{goal\_reached} claimed a win and the level completed, the
claim was right, 138 times. If it claimed a win and nothing completed,
the claim was wrong, 508 times. If it claimed nothing and the level
completed, it missed a win, 41 times. If it claimed nothing and
nothing completed, it was correctly quiet, $10{,}870$ times. The
$11{,}557$ graded actions are the runs' $11{,}614$ scored actions less
52 automatic resets, which write no log entry, and 5 entries that
share a step index. Recall is $138/179=0.771$, precision
$138/646=0.214$, and the false-alarm rate over non-winning actions
$508/11{,}378=0.045$. Most wins are claimed, and most claims are wrong.}
\label{tab:goalconf}
\begin{tabular*}{\linewidth}{@{\extracolsep{\fill}}lrrr@{}}
\toprule
 & \multicolumn{2}{c}{the action then completed\ldots} & \\
\cmidrule(lr){2-3}
the twin claimed\ldots & a level & nothing & total \\
\midrule
a win & 138 & 508 & 646 \\
nothing & 41 & $10{,}870$ & $10{,}911$ \\
\midrule
total & 179 & $11{,}378$ & $11{,}557$ \\
\bottomrule
\end{tabular*}
\end{table*}

\medskip

\begin{figure*}[t]
\centering
\includegraphics[width=\textwidth]{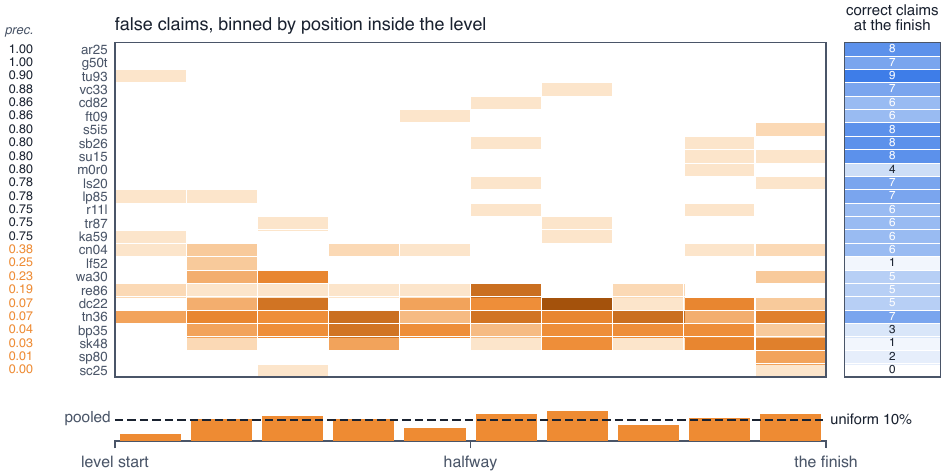}
\caption{Where a false claim lands inside its level. One row per game,
precision printed at the left. Each level's actions are cut into ten
equal parts, and orange marks the parts holding that game's false
claims, darker for more. The blue column counts correct claims; those
land at the finish by construction. The strip beneath pools all games
against the dashed uniform 10\%. The map places 334 of the 508
false claims; the remaining 174 came after a run's last completed
level, where position cannot be measured. The question is whether false
claims crowd the finish, as a predicate firing slightly early would
leave them. They do not: the median lands 0.533 through a level, and
the final tenth holds 12.6\% against the uniform 10\%.}
\label{fig:goalpred}
\end{figure*}

\paragraph{False claims land everywhere in a level.} Correct claims land at a
level's finish by construction, so the log settles a different question:
where the false ones land. Figure~\ref{fig:goalpred} shows them spread almost
evenly across the level, with no crowding near the finish. Each twin accepts
whatever its wins had in common, and those features recur throughout a level
rather than near its end.

\paragraph{Precision separates the games; recall does not.} A claim is
the predicate firing before an action, announcing that the action will
complete the level. A claim is right when the level does complete.
Recall asks whether the wins were announced, and it stays high
everywhere. Thirteen of the 25 twins claim every level they complete,
because a predicate fitted to completed levels matches the wins it has
already seen. Precision asks how often a claim is right, and it runs
from 0.00 to 1.00, failing differently at the low end in each case.
sc25 claims twice and is never right. tn36 claims 96 times and is
right 7 times, once per completed level. sk48 claims 38 times and is
right once across its 8 levels. bp35 is the one twin that gives up: it
ends the run with a predicate that returns false for every frame. The
next level loads on contact with the exit, so no single frame separates
a win from an ordinary scene. The log splits bp35's run in two,
65 false claims and then 6 completed levels with no claim at all.

\paragraph{Supervision explains the asymmetry.} Every scored action labels the
dynamics, so the log supplies $11{,}557$ dynamics examples. Completion
labels arrive once per level, so the same runs supply 179 goal examples.
That 65-to-1 ratio between the two kinds of label, rather than any
difference in the agent's effort, is what separates 79.4\% online dynamics
accuracy from 0.214 goal precision.

\paragraph{Goal precision tracks where endgame plans fail.} A false
claim carries no penalty of its own, because the harness only records
it. The cost lands on the planner, which searches for the first state
the predicate accepts, so a wrong accept ends the search early and the
plan stops short of the level. If that is the mechanism, games with low
precision should be the games whose Appendix~B endgame plans fail, and
they are. Across the 22 games with returned endgame plans, per-game
precision and the share of plans valid in the engine correlate at
$r=0.77$, Spearman $\rho=0.66$, and the three games whose endgame plans
never succeed, sp80, sk48, and tn36, hold the three lowest precisions.
Appendix~B traces sp80's five failed plans to a predicate that fires one
interact press early, and the submission log reproduces that diagnosis on its
own: five false claims, each sitting one step before a level ends. dc22
is the exception that proves the mechanism, pairing precision 0.074
with 4 valid plans out of 4. A wrong accept only truncates a plan
when it lies in the stretch the plan searches, and all 63 of dc22's
false claims land outside the last six actions of a level.

\clearpage
\twocolumn[{%
\section{Appendix K: The Learned Twins in Code}
Appendix~G reads the learning out of the twins; this appendix collects
the code, verbatim: the complete ft09 twin (Listing~\ref{lst:ft09full}),
the ka59 excerpt (Listing~\ref{lst:ka59}), and the ft09 repairs
(Listing~\ref{lst:repair}).\par\vspace{6pt}%
}]

\paragraph{A complete executable twin.} Listing~\ref{lst:ft09full}
presents the final ft09 \texttt{model.py} in full. The 305 lines hold
everything the run learned: visual parsing, action dynamics, goal
recognition, and planning. The file also keeps the quirks the agent
discovered while playing, like the animated timer row the twin excludes
from prediction. Every line and comment is the agent's, so the listing is
the model that did the planning, not a reconstruction written after the
run.

\begingroup
\captionsetup{type=listing}
\captionof{listing}{The complete ft09 twin, reproduced verbatim. The module
first describes the learned objects and constraints, then parses the board,
simulates click effects, tests whether all constraints are satisfied, and
constructs a plan to the goal. A comment identifies the animated timer as the
only region excluded from prediction. The final goal predicate is the one
confirmed through interaction. Color and line wrapping are added only for
readability. No source line is omitted or changed.}
\label{lst:ft09full}
\endgroup
\begin{lstlisting}[style=twincode,
  basicstyle=\fontsize{5.65pt}{5.92pt}\selectfont\ttfamily]
"""General twin for the click-to-copy-symbol curriculum.

The scene contains one or more overlapping 3x3 constraints.  A constraint has a
6x6 miniature key in its centre (nine 2x2 cells) and eight uniform 6x6 output
tiles on a lattice of pitch 8.  Its center is the output-colour legend for key
value 0; key value 2 denotes the other palette colour. Clicking toggles a tile.
"""
from typing import List, Dict, Any, Tuple

Grid = List[List[int]]

# The bottom row is an asynchronously advancing timer/progress animation.  It
# changes between logged game transitions as well as during them, so it cannot
# be predicted by step(grid, action).  Ignore only this isolated HUD strip.
HUD_CELLS = [(x, 63) for x in range(64)]

def _copy(g: Grid) -> Grid:
    return [r[:] for r in g]

def _uniform_block(g: Grid, x: int, y: int, n: int = 6):
    h, w = len(g), len(g[0])
    if x < 0 or y < 0 or x + n > w or y + n > h:
        return None
    v = g[y][x]
    return v if all(g[yy][xx] == v for yy in range(y, y+n)
                    for xx in range(x, x+n)) else None

def _switch_base(g: Grid, x: int, y: int):
    """Return the carrier colour of a 6x6 direction-marked switch, or None.

    A switch consists of fixed colour-6 markers over one otherwise uniform
    output-colour carrier. Ordinary live keys contain several non-6 colours,
    so cannot be confused with switches by this representation.
    """
    h, w = len(g), len(g[0])
    if not (0 <= x <= w-6 and 0 <= y <= h-6):
        return None
    vals = [g[yy][xx] for yy in range(y, y+6) for xx in range(x, x+6)]
    non6 = set(vals) - {6}
    return next(iter(non6)) if 6 in vals and len(non6) == 1 else None

def _tile_state(g: Grid, x: int, y: int):
    """State of a uniform output or colour-6-marked output switch."""
    return _uniform_block(g, x, y) if _uniform_block(g, x, y) is not None else _switch_base(g, x, y)

def _switch_offsets(g: Grid, x: int, y: int):
    """Decode the directions selected by a switch's fixed 6 markers.

    The 6x6 block has a 2x2 marker slot on each side: top=(2,0),
    left=(0,2), right=(4,2), bottom=(2,4). A full cross selects all four;
    level 5's top-only glyph selects just the output above. Self is always
    cycled separately.
    """
    if _switch_base(g, x, y) is None:
        return []
    probes = [((0,-8),(x+2,y)), ((-8,0),(x,y+2)),
              ((8,0),(x+4,y+2)), ((0,8),(x+2,y+4))]
    return [d for d,(px,py) in probes if all(g[yy][xx] == 6
            for yy in range(py,py+2) for xx in range(px,px+2))]

def _key_at(g: Grid, x: int, y: int):
    """Return a live miniature and its neighboring output values, or None."""
    h, w = len(g), len(g[0])
    if x + 6 > w or y + 6 > h:
        return None
    a = []
    for r in range(3):
        row = []
        for c in range(3):
            v = g[y+2*r][x+2*c]
            if any(g[yy][xx] != v for yy in range(y+2*r, y+2*r+2)
                   for xx in range(x+2*c, x+2*c+2)):
                return None
            row.append(v)
        a.append(row)
    outer = [a[r][c] for r in range(3) for c in range(3) if (r,c)!=(1,1)]
    # 0 and 2 are binary demands.  Later layouts use 3 as a "no cell"
    # boundary mask; 6 is reserved for an inactive/blocked miniature and may
    # not occur in a live key's outer pattern.
    if not set(outer).issubset({0, 2, 3}) or not set(outer) or 6 in outer:
        return None
    if a[1][1] in (0, 2, 3, 4, 6):
        return None
    # A key can sit on the boundary of an irregular board. Every non-masked
    # direction (outer != 3) must have a uniform output tile; masked directions
    # may be background/off-board. This subsumes the earlier full 3x3 boards.
    vals = []
    for dr in (-1, 0, 1):
        for dc in (-1, 0, 1):
            if (dr, dc) == (0, 0):
                continue
            kv = a[dr+1][dc+1]
            tx, ty = x + 8*dc, y + 8*dr
            v = _tile_state(g, tx, ty)
            # 3 alone denotes an outside-board / don't-care direction. Every
            # binary demand must point to either a plain tile or a cross switch.
            if kv != 3 and v is None:
                return None
            if kv != 3:
                vals.append(v)
    if not any(v in (0, 2) for v in outer):
        return None
    return a, vals

def _legend_palette(g: Grid) -> List[int]:
    """Read the ordered 4x4 palette swatches near the upper-right edge.

    Swatches are a contiguous vertical run but its x offset can vary as layouts
    widen. Choose the rightmost run of at least two non-background squares.
    Level 0 has no swatch, so its palette is inferred.
    """
    h, w = len(g), len(g[0])
    bg = g[0][0]
    runs = []
    for x in range(0, w-3):
        out = []
        for y in range(0, min(h-1, 32), 4):
            v = _uniform_block(g, x, y, 4)
            if v is None or v == bg:
                break
            out.append(v)
        if len(out) >= 2 and len(set(out)) == len(out):
            runs.append((x, out))
    return max(runs, default=(-1, []), key=lambda z:z[0])[1]

def _parse(g: Grid) -> Dict[str, Any]:
    h, w = len(g), len(g[0])
    constraints = []
    for y in range(h-5):
        for x in range(w-5):
            k = _key_at(g, x, y)
            if k is not None:
                key, vals = k
                constraints.append({"x": x, "y": y, "key": key,
                                    "tile_values": vals})
    # Prefer the explicit ordered swatches. Older/example-only level 0 has no
    # swatch, so infer output colours from uniform neighbors and key centres.
    palette = _legend_palette(g)
    if not palette:
        colors = set()
        for q in constraints:
            colors.update(q["tile_values"])
            colors.add(q["key"][1][1])
        palette = sorted(colors)
    return {"grid": _copy(g), "constraints": constraints, "palette": palette}

def _tile_regions(s: Dict[str, Any]):
    seen = set()
    for q in s["constraints"]:
        for dr in (-1, 0, 1):
            for dc in (-1, 0, 1):
                if ((dr, dc) == (0, 0) or q["key"][dr+1][dc+1] == 3):
                    continue
                xy = (q["x"] + 8*dc, q["y"] + 8*dr)
                if xy not in seen:
                    seen.add(xy)
                    yield xy

def _step_state(s: Dict[str, Any], action: str) -> Dict[str, Any]:
    g = _copy(s["grid"])
    if not action.startswith("ACTION6:") or len(s["palette"]) < 2:
        s["grid"] = g
        return s
    try:
        x, y = map(int, action.split(":", 1)[1].split(","))
    except (ValueError, IndexError):
        s["grid"] = g
        return s
    palette = s["palette"]
    for tx, ty in _tile_regions(s):
        if tx <= x < tx+6 and ty <= y < ty+6:
            targets = [(tx, ty)]
            # Marker position(s) say which neighboring lattice outputs this
            # switch couples to; self always cycles.
            if _switch_base(g, tx, ty) is not None:
                targets += [(tx+dx, ty+dy) for dx,dy in
                            _switch_offsets(g, tx, ty)
                            if _tile_state(g, tx+dx, ty+dy) in palette]
            for ax, ay in targets:
                old = _tile_state(g, ax, ay)
                if old not in palette:
                    continue
                new = palette[(palette.index(old) + 1) % len(palette)]
                for yy in range(ay, ay+6):
                    for xx in range(ax, ax+6):
                        if g[yy][xx] != 6:  # preserve the switch overlay
                            g[yy][xx] = new
            s["grid"] = g
            return s
    s["grid"] = g
    return s

def _render(s: Dict[str, Any]) -> Grid:
    return _copy(s["grid"])

def step(grid: Grid, action: str) -> Grid:
    return _render(_step_state(_parse(grid), action))

def _desired_color(s: Dict[str, Any], q: Dict[str, Any], key_value: int):
    """Decode a key through its center (0) and the scene's junction colour (2).

    The center explicitly labels key value 0. On multicolour levels, the one
    output colour absent from all key centers is the common junction/2 colour;
    this makes differently based boards agree on their shared 2-valued edges.
    Binary levels fall back to the colour other than this key's center.
    """
    palette = s["palette"]
    zero = q["key"][1][1]
    if len(palette) < 2 or zero not in palette:
        return None
    if key_value == 0:
        return zero
    # A binary key's 2 means the unique opposite state. With >2 states, all
    # centers collectively name the 0 states and the one unnamed palette state
    # is the shared 2/junction state.
    if len(palette) == 2:
        return next(v for v in palette if v != zero)
    centers = {qq["key"][1][1] for qq in s["constraints"]}
    junctions = [v for v in palette if v not in centers]
    return junctions[0] if len(junctions) == 1 else None

def _mismatches(s: Dict[str, Any]):
    """Unique output regions that violate any constraint."""
    bad = set()
    for q in s["constraints"]:
        for r in range(3):
            for c in range(3):
                if ((r,c) == (1,1) or q["key"][r][c] == 3):
                    continue
                tx, ty = q["x"] + 8*(c-1), q["y"] + 8*(r-1)
                want = _desired_color(s, q, q["key"][r][c])
                if _tile_state(s["grid"], tx, ty) != want:
                    bad.add((tx,ty))
    return bad

def goal_reached(grid: Grid) -> bool:
    s = _parse(grid)
    return bool(s["constraints"]) and len(s["palette"]) >= 2 and not _mismatches(s)

def plan_to_goal(grid: Grid) -> List[str] | None:
    s = _parse(grid)
    if not s["constraints"] or len(s["palette"]) < 2:
        return None
    demands: Dict[Tuple[int,int], set] = {}
    for q in s["constraints"]:
        for r in range(3):
            for c in range(3):
                if ((r,c)==(1,1) or q["key"][r][c] == 3): continue
                xy=(q["x"]+8*(c-1), q["y"]+8*(r-1))
                demands.setdefault(xy,set()).add(_desired_color(s, q, q["key"][r][c]))
    if any(len(v)>1 for v in demands.values()):
        return None

    palette = s["palette"]
    regions = sorted(set(_tile_regions(s)))
    # Cross switches couple themselves to orthogonal neighbors. Binary levels
    # have very few switches, so enumerate their click bits; all ordinary tile
    # clicks are then independently forced. This gives a true minimum plan.
    switches = [p for p in regions if _switch_base(s["grid"], *p) is not None]
    if switches and len(palette) == 2:
        start = {p: palette.index(_tile_state(s["grid"], *p)) for p in regions}
        target = dict(start)
        for p, wantset in demands.items():
            target[p] = palette.index(next(iter(wantset)))
        best = None
        for mask in range(1 << len(switches)):
            state = dict(start); actions = []
            for i, p in enumerate(switches):
                if not (mask >> i) & 1: continue
                actions.append(p)
                affected = [p] + [(p[0]+dx,p[1]+dy) for dx,dy in
                                  _switch_offsets(s["grid"], *p)]
                for q in affected:
                    if q in state: state[q] ^= 1
            for p in regions:
                if p in switches: continue
                if state[p] != target[p]:
                    state[p] ^= 1; actions.append(p)
            if state == target and (best is None or len(actions) < len(best)):
                best = actions
        return None if best is None else [f"ACTION6:{x+3},{y+3}" for x,y in best]

    plan = []
    for (x, y), wantset in sorted(demands.items(), key=lambda z:(z[0][1],z[0][0])):
        want = next(iter(wantset)); have = _tile_state(s["grid"], x, y)
        if have not in palette or want not in palette: return None
        clicks = (palette.index(want) - palette.index(have)) % len(palette)
        plan.extend([f"ACTION6:{x+3},{y+3}"] * clicks)
    return plan
\end{lstlisting}

\begingroup
\captionsetup{type=listing}
\captionof{listing}{Excerpt from the ka59 \texttt{model.py}, a learned
rule and a repaired one. \texttt{MOVES} records that each arrow moves the
token by three pixels. The rim threshold of three is the repair. A
corner-pocket counterexample left only three rim cells visible, so the
agent lowered the threshold from four. The docstrings and the
corner-pocket comment are the agent's own words. Indentation is reduced
for width, and pixel loops are shown through the named helpers
\texttt{interior} and \texttt{eight\_neighbours}.}
\label{lst:ka59}
\endgroup
\begin{lstlisting}[style=twincode]
"""Observed world model for the unknown
ARC game. Until an action effect has
been observed, it is deliberately
represented as a no-op. Rules are added
only from bridge transitions and must
keep validating against the complete
transition log."""

MOVES = {
    "ACTION1": (0, -3),
    "ACTION2": (0, 3),
    "ACTION3": (-3, 0),
    "ACTION4": (3, 0),
}

def _active_center(grid):
    """The movable token has a color-0
    center and a color-14 rim."""
    # loops abbreviated
    for x, y in interior(grid):
        rim = eight_neighbours(grid, x, y)
        # A token inside a plus's 3x3 corner
        # pocket has connector zeros on two
        # adjacent edges, leaving only three
        # visible rim cells.
        if (all(v in (0, 14) for v in rim)
                and rim.count(14) >= 3):
            return x, y
\end{lstlisting}

\begingroup
\captionsetup{type=listing}
\captionof{listing}{Two repairs to the ft09 twin, in run order. The first
pair shows the dynamics rule before and after the counterexample at
scored action 0, which exposed two wrong cells in the bottom
action-budget bar. The second pair shows the goal predicate before and
after the level-0 boundary at scored action 3. The original predicate
matches one finished board and accepts the unsolved level-1 board; the
rewrite states a constraint and rejects it. Comments are the agent's own
words. The excerpts reduce indentation for width, and the repaired
dynamics block shows only the newly added lines.}
\label{lst:repair}
\endgroup
\textbf{\footnotesize Dynamics, at scored action 0}\par\vspace{2pt}
\begin{lstlisting}[style=twincode]
old = g[sy[r]][sx[c]]
if old in (8, 9):
  new = 17 - old
  for yy in range(sy[r], sy[r] + th):
    for xx in range(sx[c], sx[c] + tw):
      g[yy][xx] = new
state["grid"] = g
\end{lstlisting}
\vspace{1pt}
\textbf{\footnotesize Dynamics, after its counterexample}\par\vspace{2pt}
\begin{lstlisting}[style=twincode]
  # Bottom HUD is an action budget: every effective click
  # consumes two cyan(c) cells, filled dark-cyan(b) right-to-left.
  available = [xx for xx, v in enumerate(g[-1]) if v == 12]
  for xx in available[-2:]:
    g[-1][xx] = 11
\end{lstlisting}
\vspace{1pt}
\textbf{\footnotesize Goal, before the level-0 boundary}\par\vspace{2pt}
\begin{lstlisting}[style=twincode]
def goal_reached(grid: Grid) -> bool:
  s = _parse(grid)
  want = _desired(s)
  if want is None:
    return False
  return all(
    (r, c) == (1, 1) or s["tiles"][r][c] == want[r][c]
    for r in range(3) for c in range(3)
  )
\end{lstlisting}
\vspace{1pt}
\textbf{\footnotesize Goal, after it}\par\vspace{2pt}
\begin{lstlisting}[style=twincode]
def goal_reached(grid: Grid) -> bool:
  s = _parse(grid)
  return (bool(s["constraints"])
          and len(s["palette"]) == 2
          and not _mismatches(s))
\end{lstlisting}

\end{document}

%% file: compute_pergame.tex
\begin{tabular}{l r r r r r}
\toprule
game & proc & n-c & h & act & tok/act \\
\midrule
ar25 & 24.5 & 0.80 & 1.2 & 254 & 96k \\
bp35 & 161.8 & 3.56 & 6.5 & 529 & 306k \\
cd82 & 34.9 & 0.83 & 1.6 & 93 & 375k \\
cn04 & 97.3 & 2.02 & 3.9 & 261 & 373k \\
dc22 & 269.4 & 4.15 & 7.8 & 1,219 & 221k \\
ft09 & 16.8 & 0.48 & 1.3 & 80 & 210k \\
g50t & 115.5 & 2.51 & 4.2 & 577 & 200k \\
ka59 & 625.3 & 9.59 & 17.8 & 1,138 & 549k \\
lf52 & 46.1 & 0.63 & 1.5 & 829 & 56k \\
lp85 & 27.0 & 0.59 & 0.9 & 104 & 260k \\
ls20 & 191.9 & 3.77 & 7.3 & 515 & 373k \\
m0r0 & 77.3 & 1.09 & 1.8 & 236 & 328k \\
r11l & 34.5 & 0.94 & 1.3 & 85 & 406k \\
re86 & 98.7 & 1.94 & 3.5 & 750 & 132k \\
s5i5 & 94.0 & 2.27 & 3.0 & 348 & 270k \\
sb26 & 5.1 & 0.25 & 0.7 & 137 & 37k \\
sc25 & 18.8 & 0.48 & 3.7 & 701 & 27k \\
sk48 & 145.2 & 2.55 & 3.6 & 752 & 193k \\
sp80 & 206.6 & 2.61 & 6.1 & 1,037 & 199k \\
su15 & 48.5 & 1.24 & 2.3 & 129 & 376k \\
tn36 & 102.8 & 1.43 & 2.7 & 475 & 216k \\
tr87 & 61.2 & 1.26 & 3.9 & 162 & 377k \\
tu93 & 61.4 & 1.50 & 2.1 & 236 & 260k \\
vc33 & 11.3 & 0.67 & 1.5 & 243 & 47k \\
wa30 & 24.4 & 0.68 & 1.2 & 724 & 34k \\
\midrule
\textbf{total} & \textbf{2.60\,B} & \textbf{47.8} & \textbf{91.4} & \textbf{11,614} & \textbf{224k} \\
\bottomrule
\end{tabular}

%% file: perlevel_a.tex
\begin{tabular}{llrrrrrrrrrr}
\toprule
Game & System & L1 & L2 & L3 & L4 & L5 & L6 & L7 & L8 & L9 & L10 \\
\midrule
tu93 & Human & 19 & 16 & 34 & 42 & 123 & 80 & 14 & 23 & 111 &  \\
 & EWM & \textbf{18} & 16 & \textbf{19} & \textbf{18} & \textbf{29} & \textbf{28} & \textbf{14} & \textbf{21} & \textbf{29} &  \\
 & OPINE-World & 22 & 32 & \textbf{19} & 37 & 31 & 36 & \textbf{14} & 25 & 55 &  \\
 & \twin{} & \textbf{18} & \textbf{15} & \textbf{19} & \textbf{18} & 37 & 40 & 16 & \textbf{21} & 52 &  \\
\addlinespace[2.2pt]
sb26 & Human & 18 & 28 & 18 & 19 & 31 & 23 & 58 & 18 &  &  \\
 & EWM & 13 & \textbf{15} & \textbf{15} & \textbf{15} & \textbf{17} & \textbf{19} & 39 & \textbf{17} &  &  \\
 & OPINE-World & 13 & 47 & 31 & \textbf{15} & \textbf{17} & 39 & 35 & \textbf{17} &  &  \\
 & \twin{} & \textbf{9} & 28 & \textbf{15} & \textbf{15} & \textbf{17} & \textbf{19} & \textbf{17} & \textbf{17} &  &  \\
\addlinespace[2.2pt]
lp85 & Human & 17 & 38 & 31 & 16 & 41 & 60 & 26 & 159 &  &  \\
 & EWM & 14 & 138 & 19 & \textbf{13} & \textbf{10} & 20 & \textbf{5} & \textbf{7} &  &  \\
 & OPINE-World & 8 & 12 & 17 & 16 & 11 & 20 & 8 & 17 &  &  \\
 & \twin{} & \textbf{5} & \textbf{8} & \textbf{16} & \textbf{13} & 11 & \textbf{19} & 8 & 24 &  &  \\
\addlinespace[2.2pt]
ar25 & Human & 32 & 50 & 75 & 37 & 89 & 159 & 233 & 73 &  &  \\
 & EWM & 17 & 14 & 41 & \textbf{22} & 30 & 56 & \textbf{37} & \textbf{47} &  &  \\
 & OPINE-World & 17 & 16 & 75 & 24 & 34 & 55 & 112 & \textbf{47} &  &  \\
 & \twin{} & \textbf{15} & \textbf{12} & \textbf{40} & \textbf{22} & \textbf{28} & \textbf{53} & \textbf{37} & \textbf{47} &  &  \\
\addlinespace[2.2pt]
tr87 & Human & 54 & 58 & 40 & 45 & 71 & 146 &  &  &  &  \\
 & EWM & 47 & 190 & 47 & 36 & 47 & 173 &  &  &  &  \\
 & OPINE-World & 33 & 29 & \textbf{26} & 27 & \textbf{22} & 74 &  &  &  &  \\
 & \twin{} & \textbf{21} & \textbf{26} & \textbf{26} & \textbf{21} & 45 & \textbf{23} &  &  &  &  \\
\addlinespace[2.2pt]
r11l & Human & 22 & 33 & 51 & 26 & 52 & 49 &  &  &  &  \\
 & EWM & \textbf{5} & 22 & 20 & \textbf{13} & 90 & 77 &  &  &  &  \\
 & OPINE-World & 7 & 13 & 35 & 16 & 23 & 33 &  &  &  &  \\
 & \twin{} & \textbf{5} & \textbf{11} & \textbf{17} & \textbf{13} & \textbf{19} & \textbf{20} &  &  &  &  \\
\addlinespace[2.2pt]
ft09 & Human & 43 & 12 & 23 & 28 & 65 & 37 &  &  &  &  \\
 & EWM & \textbf{4} & \textbf{7} & \textbf{14} & 86 & 23 & 23 &  &  &  &  \\
 & OPINE-World & 6 & \textbf{7} & \textbf{14} & \textbf{16} & 55 & \textbf{13} &  &  &  &  \\
 & \twin{} & \textbf{4} & \textbf{7} & \textbf{14} & 21 & \textbf{21} & \textbf{13} &  &  &  &  \\
\addlinespace[2.2pt]
cd82 & Human & 55 & 8 & 41 & 21 & 23 & 23 &  &  &  &  \\
 & EWM & 16 & \textbf{6} & 65 & \textbf{14} & \textbf{13} & \textbf{16} &  &  &  &  \\
 & OPINE-World & 74 & \textbf{6} & \textbf{20} & 23 & 20 & 17 &  &  &  &  \\
 & \twin{} & \textbf{13} & \textbf{6} & 25 & 20 & \textbf{13} & \textbf{16} &  &  &  &  \\
\addlinespace[2.2pt]
cn04 & Human & 29 & 54 & 85 & 300 & 208 & 113 &  &  &  &  \\
 & EWM & \textbf{14} & 199 & \textbf{22} & \textbf{29} & 124 & 60 &  &  &  &  \\
 & OPINE-World & 32 & 50 & 32 & 39 & \textbf{49} & 61 &  &  &  &  \\
 & \twin{} & 20 & \textbf{47} & 54 & \textbf{29} & 70 & \textbf{41} &  &  &  &  \\
\addlinespace[2.2pt]
su15 & Human & 22 & 42 & 26 & 115 & 36 & 31 & 8 & 40 & 41 &  \\
 & EWM & 18 & 89 & 19 & \textbf{14} & 7 & 58 & \textbf{6} & 54 & \textbf{19} &  \\
 & OPINE-World & 23 & 37 & \textbf{16} & 99 & 24 & \textbf{18} & 7 & 59 & 51 &  \\
 & \twin{} & \textbf{12} & \textbf{16} & 18 & 21 & \textbf{5} & 22 & \textbf{6} & \textbf{8} & 21 &  \\
\addlinespace[2.2pt]
re86 & Human & 26 & 42 & 86 & 108 & 189 & 139 & 424 & 241 &  &  \\
 & EWM & 23 & 38 & 124 & \textbf{64} & 1,505 &  &  &  &  &  \\
 & OPINE-World & 21 & \textbf{36} & 52 & 65 & 71 & \textbf{70} & 214 & 321 &  &  \\
 & \twin{} & \textbf{20} & \textbf{36} & \textbf{47} & 247 & \textbf{63} & 74 & \textbf{108} & \textbf{155} &  &  \\
\addlinespace[2.2pt]
tn36 & Human & 32 & 72 & 26 & 40 & 30 & 55 & 62 &  &  &  \\
 & EWM & 11 & 22 & 14 & 137 & 264 & 1,012 & 1,598 &  &  &  \\
 & OPINE-World & 13 & 23 & \textbf{9} & 111 & 136 & 62 & \textbf{63} &  &  &  \\
 & \twin{} & \textbf{7} & \textbf{11} & 166 & \textbf{29} & \textbf{35} & \textbf{45} & 182 &  &  &  \\
\addlinespace[2.2pt]
vc33 & Human & 7 & 18 & 44 & 61 & 131 & 34 & 152 &  &  &  \\
 & EWM & \textbf{3} & 10 & 54 & 1,529 &  &  &  &  &  &  \\
 & OPINE-World & 10 & 11 & 28 & 175 & 91 & 36 & 75 &  &  &  \\
 & \twin{} & 4 & \textbf{8} & \textbf{24} & \textbf{47} & \textbf{89} & \textbf{20} & \textbf{51} &  &  &  \\
\bottomrule
\end{tabular}

%% file: perlevel_b.tex
\begin{tabular}{llrrrrrrrrrr}
\toprule
Game & System & L1 & L2 & L3 & L4 & L5 & L6 & L7 & L8 & L9 & L10 \\
\midrule
m0r0 & Human & 30 & 111 & 203 & 26 & 500 & 237 &  &  &  &  \\
 & EWM & 20 & 534 & 79 & 13 & 408 & 1,519 &  &  &  &  \\
 & OPINE-World & \textbf{19} & \textbf{35} & \textbf{76} & 16 & 56 & 57 &  &  &  &  \\
 & \twin{} & \textbf{19} & \textbf{35} & 89 & \textbf{11} & \textbf{42} & \textbf{40} &  &  &  &  \\
\addlinespace[2.2pt]
sc25 & Human & 36 & 6 & 32 & 83 & 143 & 50 &  &  &  &  \\
 & EWM & 20 & \textbf{5} & 63 & 1,019 &  &  &  &  &  &  \\
 & OPINE-World & 32 & \textbf{5} & 36 & \textbf{32} & \textbf{50} & \textbf{101} &  &  &  &  \\
 & \twin{} & \textbf{19} & \textbf{5} & \textbf{30} & 647 &  &  &  &  &  &  \\
\addlinespace[2.2pt]
sp80 & Human & 39 & 58 & 25 & 148 & 96 & 152 &  &  &  &  \\
 & EWM & \textbf{6} & 780 &  &  &  &  &  &  &  &  \\
 & OPINE-World & 10 & 30 & 56 & 43 & 85 & \textbf{145} &  &  &  &  \\
 & \twin{} & 11 & \textbf{15} & \textbf{12} & \textbf{40} & \textbf{31} & 928 &  &  &  &  \\
\addlinespace[2.2pt]
wa30 & Human & 71 & 119 & 183 & 98 & 368 & 68 & 79 & 442 & 415 &  \\
 & EWM & 429 & 121 & 181 & 97 & 233 & \textbf{48} & \textbf{42} & 139 & 205 &  \\
 & OPINE-World & 49 & 169 & 80 & 71 & 245 & 54 & 53 & 440 & 304 &  \\
 & \twin{} & \textbf{26} & \textbf{48} & \textbf{76} & \textbf{55} & \textbf{124} & 138 & 60 & \textbf{130} & \textbf{67} &  \\
\addlinespace[2.2pt]
g50t & Human & 78 & 175 & 179 & 230 & 96 & 54 & 67 &  &  &  \\
 & EWM & \textbf{58} & 138 & 85 & \textbf{99} & \textbf{55} & \textbf{48} & \textbf{71} &  &  &  \\
 & OPINE-World & 90 & 78 & \textbf{75} & 134 & 158 & 122 & 100 &  &  &  \\
 & \twin{} & 70 & \textbf{44} & 83 & 127 & 104 & 54 & 95 &  &  &  \\
\addlinespace[2.2pt]
ls20 & Human & 22 & 123 & 73 & 84 & 96 & 192 & 186 &  &  &  \\
 & EWM & 22 & 97 & 74 & 101 & 76 & 216 & 128 &  &  &  \\
 & OPINE-World & 17 & \textbf{75} & 103 & 93 & 129 & 359 & 183 &  &  &  \\
 & \twin{} & \textbf{16} & 101 & \textbf{43} & \textbf{49} & \textbf{74} & \textbf{113} & \textbf{119} &  &  &  \\
\addlinespace[2.2pt]
ka59 & Human & 28 & 109 & 51 & 51 & 33 & 132 & 326 &  &  &  \\
 & EWM & 112 & 66 & \textbf{35} & 61 & 93 & 552 & \textbf{180} &  &  &  \\
 & OPINE-World & 45 & 76 & 66 & 165 & 27 & \textbf{104} & 593 &  &  &  \\
 & \twin{} & \textbf{40} & \textbf{45} & 134 & \textbf{56} & \textbf{20} & 124 & 719 &  &  &  \\
\addlinespace[2.2pt]
dc22 & Human & 59 & 102 & 67 & 98 & 324 & 578 &  &  &  &  \\
 & EWM & 66 & \textbf{46} & 92 & 114 & 1,524 &  &  &  &  &  \\
 & OPINE-World & 132 & 74 & 53 & 120 & \textbf{125} & 758 &  &  &  &  \\
 & \twin{} & \textbf{26} & 69 & \textbf{49} & \textbf{64} & 444 & \textbf{567} &  &  &  &  \\
\addlinespace[2.2pt]
sk48 & Human & 61 & 177 & 101 & 103 & 230 & 181 & 125 & 92 &  &  \\
 & EWM & \textbf{14} & 411 & 83 & 558 & 76 & 1,681 &  &  &  &  \\
 & OPINE-World & 18 & 116 & 64 & 236 & \textbf{66} &  &  &  &  &  \\
 & \twin{} & 24 & \textbf{68} & \textbf{41} & \textbf{59} & 112 & \textbf{123} & \textbf{146} & \textbf{179} &  &  \\
\addlinespace[2.2pt]
lf52 & Human & 32 & 81 & 60 & 71 & 205 & 148 & 244 & 109 & 164 & 225 \\
 & EWM & \textbf{8} & 209 & 81 & 58 & \textbf{91} & 151 & 2,048 &  &  &  \\
 & OPINE-World & 11 & 267 & 104 & 118 &  &  &  &  &  &  \\
 & \twin{} & \textbf{8} & \textbf{34} & \textbf{46} & \textbf{50} & 124 & \textbf{93} & \textbf{179} & \textbf{79} & \textbf{112} & \textbf{104} \\
\addlinespace[2.2pt]
bp35 & Human & 21 & 48 & 44 & 38 & 33 & 87 & 86 & 131 & 163 &  \\
 & EWM & 20 & 183 & 53 & \textbf{36} & 50 &  &  &  &  &  \\
 & OPINE-World & 19 & 343 & 141 &  &  &  &  &  &  &  \\
 & \twin{} & \textbf{17} & \textbf{79} & \textbf{36} & 64 & \textbf{33} & \textbf{66} & \textbf{64} & \textbf{45} & \textbf{125} &  \\
\addlinespace[2.2pt]
s5i5 & Human & 20 & 89 & 106 & 54 & 162 & 38 & 86 & 83 &  &  \\
 & EWM & 170 & 948 & 396 & 1,514 &  &  &  &  &  &  \\
 & OPINE-World & 23 & 74 & 125 & \textbf{40} & 376 &  &  &  &  &  \\
 & \twin{} & \textbf{15} & \textbf{35} & \textbf{57} & 43 & \textbf{40} & \textbf{29} & \textbf{87} & \textbf{42} &  &  \\
\bottomrule
\end{tabular}